\documentclass[letterpaper, 10 pt, conference]{ieeeconf}  

\usepackage{amsmath,amsfonts,amssymb}
\usepackage{algorithm}
\usepackage{float}
\newfloat{algorithm}{t}{lop}
\usepackage{xcolor}
\usepackage{algorithmic}
\usepackage{array}
\usepackage{textcomp}
\usepackage{stfloats}
\usepackage{url}
\usepackage{verbatim}
\usepackage{graphicx}
\usepackage{cite}
\usepackage{booktabs}
\usepackage{todonotes}
\usepackage{multirow}
\usepackage[caption=false,font=footnotesize]{subfig}
\usepackage{fixltx2e}

\definecolor{highlander_blue}{RGB}{0,61,165}

\makeatletter
\newcommand\fs@spaceruled{\def\@fs@cfont{\bfseries}\let\@fs@capt\floatc@ruled
  \def\@fs@pre{\vspace{0.5\baselineskip}\hrule height.8pt depth0pt \kern2pt}%
  \def\@fs@post{\kern2pt\hrule\vspace{-0.95\baselineskip}}
  \def\@fs@mid{\kern2pt\hrule\kern2pt}%
  \let\@fs@iftopcapt\iftrue}
\makeatother

\IEEEoverridecommandlockouts                              

\title{\LARGE \bf
A Koopman Operator-based NMPC Framework for \\Mobile Robot Navigation under Uncertainty 
}

\author{Xiaobin Zhang,$^{1,*}$ Mohamed Karim Bouafoura,$^{2,*}$ Lu Shi,$^{3}$ and Konstantinos Karydis$^{1}$
\thanks{
\noindent $^{*}$ Equal Contribution.
$^{1}$ Department of Electrical and Computer Engineering, University of California, Riverside, 900 University Avenue, Riverside, CA 92521, USA. Email: \{xzhan548, karydis\}@ucr.edu. 
$^{2}$ Polytechnic School of Tunisia, BP 743, 2078 La Marsa, Tunisia,
email: {mohamed.bouafoura@ept.rnu.tn}. 
$^{3}$ Institute of AI Industry Research, Tsinghua University, China, email: {shilu@air.tsinghua.edu.cn}.} 
\thanks{
\noindent We gratefully acknowledge the support of NSF \#IIS-1910087, \#CMMI-2046270, and \#CNS-2312395. Any opinions, findings, and conclusions or recommendations expressed in this material are those of the authors and do not necessarily reflect the views of the National Science Foundation.}
}

\begin{document}

\maketitle
\thispagestyle{empty}
\pagestyle{empty}

\begin{abstract}

Mobile robot navigation can be challenged by system uncertainty.  For example, ground friction may vary abruptly causing slipping, and noisy sensor data can lead to inaccurate feedback control.  Traditional model-based methods may be limited when considering such variations, making them fragile to varying types of uncertainty.  One way to address this is by leveraging learned prediction models by means of the Koopman operator into nonlinear model predictive control (NMPC).  This paper describes the formulation of, and provides the solution to, an NMPC problem using a lifted bilinear model that can accurately predict affine input systems with stochastic perturbations.  System constraints are defined in the Koopman space, while the optimization problem is solved in the state space to reduce computational complexity.  Training data to estimate the Koopman operator for the system are given via randomized control inputs.  The output of the developed method enables closed-loop navigation control over environments populated with obstacles.  The effectiveness of the proposed method has been tested through numerical simulations using a wheeled robot with additive stochastic velocity perturbations, Gazebo simulations with a realistic digital twin robot, and physical hardware experiments without knowledge of the true dynamics.
\end{abstract}



\section{Introduction}

The presence of uncertainty can challenge mobile robot navigation in practical deployment. 
For example, different ground surfaces can have varying levels of friction. These variations can cause the robot to slip or skid, especially when suddenly turning or stopping. 
This makes it difficult for the robot to maintain precise control and follow its planned path accurately~\cite{wang2008modeling}.  
Further, noise in sensor data can lead to inaccurate control, and varying environmental conditions, such as in lighting, can affect sensor readings and the robot's ability to perceive its surroundings accurately~\cite{ji2022proactive}. 

Model-based control techniques such as Model Predictive Control (MPC) and variants built on top of MPC have been extensively used to drive mobile robots. 
Examples include long trajectory tracking to follow complex and extended paths precisely~\cite{li2015trajectoryMPC} and accurate local-path tracking~\cite{pacheco2015MPC}. 
However, disturbances along the process can degrade modeling accuracy and the resulting recursive feasibility and stability of an MPC algorithm~\cite{grune2017nonlinear}. 
To address this, robust MPC~\cite{fleming2014robust,limon2006input} considers the worst case, i.e. assumes a maximum bound on the disturbance, while stochastic MPC constitutes a good alternative to handle stochastic disturbances~\cite{sun2019self}. 
%
However, generalizing model-based control may be challenged as the operating environments become more diverse, thus necessitating additional tuning before deployment~\cite{edwards2021automatic}, while the underlying models may become invalid as new unmodeled uncertainties may arise during deployment~\cite{karydis2015probabilistically}. 

Data-driven control has emerged as a generalizable means to handle runtime uncertainties. 
Deep learning methods leverage vast amounts of data to create models that can predict and adjust to various operational conditions~\cite{karoly2020deep}. 
Reinforcement Learning, in particular, enables robots to learn optimal control policies through trial and error, improving their performance in dynamic and uncertain environments~\cite{oikawa2021reinforcement}. 
With the recent development of Large Language Models, the decision-making processes can be enhanced by interpreting complex instructions and adapting to nuanced changes in the environment~\cite{chen2024llm}. 
%
Despite the demonstrated strengths of those methods, they still lack in their capacity to be deployed on edge devices mounted on robots while requiring vast datasets for training~\cite{zhao2024deep}, which may not even be available (yet) in some applications.

More recently, Koopman operator theory has been gaining attention as a means to enable real-time learning and support data-driven control~\cite{shi2024koopman}. 
The Koopman operator is a powerful tool for learning nonlinear dynamics by lifting the variables in the state space into observables in the Koopman space, where the dynamics is considered to be linear but infinite-dimensional. 
It is not nearly as data-intensive as most other machine learning techniques and takes much less time to train and calculate during deployment (as shown for instance in~\cite{zhou2025learning} to estimate a parallel mechanism's kinematics), thus making it a suitable candidate method for hardware with lower computation capabilities and requiring a higher control frequency. 
The Koopman operator, along with estimation methods such as the Extended Dynamic Mode Decomposition with control (EDMDc)~\cite{korda2018linear}, has been widely applied in the modeling and control of various robotic systems~\cite{shi2023koopman}. 
Notable examples include the modeling and control of a tail-actuated robotic fish~\cite{mamakoukas2021Taylor}, trajectory control for micro-aerial vehicles~\cite{shi2020data}, dynamics estimation for spherical robots~\cite{abraham2017RKexample}, and model extraction for soft systems~\cite{bruder2019ICRA}.

This work aims to integrate Koopman-based model learning into a nonlinear MPC (NMPC) framework for mobile robot navigation in obstacle-cluttered environments under uncertainty, such as stochastic velocity perturbations caused by wheel slipping. 
We extend EDMDc for a bilinear system framework, where a least squares problem based on the Kronecker product can reduce the parameter synthesis compared to solving the equivalent nonlinear optimization. 
%
Contrary to existing Koopman-based MPC/NMPC methods that transpose the whole problem to the lifted space, we formulate the control problem in the state space by setting appropriate constraints. 
This improves optimization complexity and reduces the required computational time essential for real-time implementation.
  
The method is first tested in simulation with a differential drive robot model with stochastic control velocity perturbations. 
Then, it is validated in Gazebo simulation with a wheeled mobile robot with \emph{unknown dynamics that are learned in real-time via the Koopman operator}. 
Further, physical hardware experiments demonstrate the practical feasibility of the developed method. 
To the best of our knowledge, this is the first work to utilize a Kronecker product-based bilinear Koopman model for learning-based navigation control of physical mobile robots. 

\section{Technical Background}\label{2}
Consider the control-affine dynamical system of the form
\begin{equation}\label{eq:affine}
{\mathop x\limits^. (t) = f(x(t)) + \sum_{i=1}^mg_i(x(t ))u_i(t)+d(x,u,t)}\;,
\end{equation}
where $x(t)=[x_1(t),...,x_n(t)]^T \in X \subset \mathbb{R}^n$ is the state at time $t$, and $u(t)=[u_1(t),...,u_m(t)]^T \in X \subset \mathbb{R}^n$ is the input at time $t \in \mathbb{N}$.
Functions $f$ and $g$ denote the drift and control vector fields respectively, while $d \in \mathbb{R}^{n}$  encompasses the overall model uncertainty, i.e. parameter uncertainty as well as unknown disturbances. 
The control-affine system~\eqref{eq:affine} can represent the dynamics of several types of mobile robots~\cite{zhou2023safe}. 

%


\subsection{Koopman Generators of Control-Affine Systems} \label{3}
\subsubsection{Primer on Koopman Operator Theory}
First, consider the unforced dynamical system 
$\dot{x}=f(x)$ 
and let $\Phi(t,x)$ be its flow map. 
Let $\mathcal{F}$ be the space of all complex-valued observables denoted $\varphi$. 
The continuous time Koopman operator $\mathcal{K}:\mathcal{F}\to  \mathcal{F}$ is defined as 
$(\mathcal{K}\varphi)(\cdot)=\varphi \circ \Phi(t,\cdot)$. 
Being linear, the Koopman operator can be characterized by its eigenvalues and eigenfunctions. A function $\phi$ is an eigenfunction of $\mathcal{K}$ if  
$(\mathcal{K}\phi)(\cdot)=e^{\lambda t}\phi(\cdot)$, 
with $\lambda$ being the corresponding eigenvalue~\cite{nathan2018applied}. 
%
The time-varying observable $ \psi(t, x) = \mathcal{K}\phi(x)$ is the solution of the PDE~\cite{surana2016koopman}
\begin{equation}\label{e8}
\begin{array}{c}
\frac{\partial \psi}{\partial t}= \mathcal{L}_f \psi \\
\psi(0, x) = \phi(x_0)
\end{array}\;,
\end{equation}
where $x_0$ is the initial condition for the unforced system and $\mathcal{L}_f$ is the Lie derivative with respect to $f$.

A vector-valued observable $\mathrm{g}(\cdot)$ may be expressed in terms of Koopman eigenfunctions $\phi_i$ as 
$\mathrm{g}(\cdot) = \sum_{i=1}^{\infty }\phi_i(\cdot)\mathrm{v}_i$, 
where $\mathrm{v}_i$ denote 
the Koopman modes. 
Koopman modes are obtained from the projection of the observable on the span of Koopman eigenfunctions. 
Then, 
\begin{equation*}\label{e10}
\mathcal{K}\mathrm{g}(\cdot) =\mathcal{K} \sum_{i=1}^{\infty }\phi_i(\cdot)\mathrm{v}_i 
=  \sum_{i=1}^{\infty } \mathcal{K}\phi_i(\cdot)\mathrm{v}_i
= \sum_{i=1}^{\infty } \lambda_i\phi_i(\cdot)\mathrm{v}_i\;.
\end{equation*}
It is worth noting that Koopman eigenvalues and eigenfunctions are properties of the dynamics only, while Koopman modes depend on the observable. 

\subsubsection{Bilinear Koopman Models}
The canonical Koopman model described in~\ref{3} is a linear one, describing the evolution of the observable in a linear manner. However, in most cases with control systems, the system dynamics models are nonlinear, where the linear realization of the Koopman operator may not represent advantages in terms of computational efficiency. 
Certain types of realizations are particularly amenable to control design since the control input appears either linearly or bilinearly in them~\cite{bruder2021advantages}. Therefore, in this work we consider a bilinear Koopman model for better approximation of the true dynamics model.


A Koopman-based model realization of~\eqref{eq:affine} over a set of observables $\{z_i \in Z\}_{i=1}^N $  is bilinear if there exist sets of coefficients $\{ a_{ij} \in \mathbb{R} \}_{i=1,j=1}^{N,N}$, $\{ b_{ij} \in \mathbb{R} \}_{i=1,j=1}^{N,m}$, $\{ h_{ijk} \in \mathbb{R} \}_{i=1,j=1,k=1}^{N,N,m}$, and $\{ c_{ij} \in \mathbb{R} \}_{i=1,j=1}^{n,N}$ such that~\cite[Definition II.1]{bruder2021advantages}
\begingroup\makeatletter\def\f@size{8.5}\check@mathfonts
\def\maketag@@@#1{\hbox{\m@th\large\normalfont#1}}
\begin{equation}\label{eq:bilinear_model_definition}
    \frac{d}{dt} z_i(t)= \sum_{j=1}^N a_{ij}z_j(t) + \sum_{j=1}^m b_{ij}u_j(t) + \sum_{j=1}^m  \sum_{jk=1}^N h_{ijk}z_k(t)u_j(t)
\end{equation}
\endgroup
$\exists h_{ijk} \neq 0$, for $i=1,...,N$ and
\begin{equation}\label{eq:next_state_prediction}
    x_i(t)=\sum_{j=1}^N c_{ij}z_j(t)
\end{equation}
for $i=1,...,n$. 
For a control-affine system, if a set of observables $\Bar{Z} = \{z_i \in Z\}_{i=1}^N $ is a basis of $Z$,
then the realization of the system defined over $\Bar{Z}$ is bilinear~\cite[Corollary II.1]{bruder2021advantages}.

This way, ~\eqref{eq:bilinear_model_definition} and ~\eqref{eq:next_state_prediction} can be used to predict the dynamics of $x(t)$ along new trajectories given its history. 
State $x$ is first projected into observable $z$ in the Koopman space as $z=\varphi(x)$ via the lifting function $\varphi(x,u)=[\varphi_1(x,u),\ldots,\varphi_M(x,u)]$, where $\{\varphi_i:X\times U \rightarrow \mathbb{R}\}^M_{i=1}$ constitute a finite-dimensional subspace as the span
of a user-defined set of $M$ linearly independent observables; a systematic way to select the lifting function is discussed in~\cite{shi2021acd}. 
Then, matrices $A$, $B$, and $H$, as well as the control vector $u$, are governing the evolution of the observable. 
Finally, the evolved state $x$ is obtained by projecting the evolved observable back to the Cartesian space by matrix $C$. 

\subsection{Kronecker Product-based Formulation} \label{4}
As a direct extension of~\cite{bichiou2018time}, the bilinear system~\eqref{eq:bilinear_model_definition} can be expressed using the Kronecker product notation as
%
\begin{equation}\label{eq:bilinear_actual}
\dot{z}(t)= Az(t)+Bu(t)+H(u(t)\otimes z(t))\;. 
\end{equation}
Given a collection of sampled time histories of inputs $U=[u_1,\cdots,u_{M-1}]$ and observations $Z=[z_1,\cdots,z_{M-1}]$, matrices $A$, $B$, and $H$ can be obtained as the best bilinear predictor in the lifted space in a least-squares sense, that is 
$\underset{A,B,H}{min}\left\| Z'-AZ-BU-H(U\otimes Z)\right\|$, 
where $Z'=[z_2,\cdots,z_{M}]$. 
Matrix $C$ is obtained in a linear least-squares sense as
$\underset{C}{min}\left\| X-CZ\right\|$, 
where $X=[x_1,\cdots,x_{M-1}]$ are the measurements in the state space. 
The analytical solutions of these least-squares problems are 
\begin{equation}\label{eq:koopman_sol}
\left\{\begin{array}{l}
[A,B,H]=Z'[Z,U\otimes Z,U]^+ ~~\textrm{and}\\
C=XZ^+
\end{array}\right.\;,
\end{equation}
respectively; $(\cdot)^+$ denotes the Moore–Penrose pseudoinverse. 

\section{Proposed Koopman-NMPC Formulation}
We are now ready to present the main contribution of this work. 
The goal is for a wheeled mobile robot with uncertainty affecting its motion to reach a desired pose while avoiding collisions. 
There is no prior knowledge about the nonlinear dynamics model of the robot. 
%
We seek to solve 
%
\begin{equation}\label{eq:optim}
\begin{array}{l}
\underset{U_k,X_k}{min}\quad  J_K (x,u)=\sum_{k=0}^{K-1} \ell(x_k,u_k)\\
\textrm{subject to:} \\
\hspace{2cm}
\begin{array}{l}
f_{eq}(X_k,U_k)=0\\
x(0)=x_0\\
u_k \in \mathbb{U} \quad  \forall k \in [0,K-1]\\
x_k \in \mathbb{X} \quad  \forall k \in [0,K]\\
-\left\| x_k-x_{oi}\right\|+r+r_{oi}\leqslant 0
\end{array}
\end{array}
\end{equation}
%
where $x_{oi}$ and $r_{oi}$ represent the coordinates and radius of the $i$-th obctacle, respectively, and $r$ is the controlled robot radius. 
In problem~\eqref{eq:optim}, $f_{eq}$ contains the dynamic constraints along the prediction horizon $K$. 
These are given by
\begin{equation}\label{eq11}
f_{eq}=\begin{bmatrix} x_{k|k}-x_k\\
x_{k+1|k}-f_b(x_{k|k},u_{k}) \\
 \vdots \\
x_{k+K|k}-f_b(x_{k+K-1|k},u_{k+K-1}) \\
\end{bmatrix}\;,
\end{equation}
in which $x_{k+i|k}$ and $u_{k+i}$ denote the state at time instant $k+i$ predicted at time instant $k$ and the control input at time $k+i$, respectively; $x_k$ here refers to the current state. 
The Koopman-bilinear prediction handling system uncertainty, $f_b(\cdot)$, introduced in equality constraints~\eqref{eq11} is given by 
\begin{equation}\label{eq12}
\begin{array}{ll}
f_b(x_{k+i|k},u_{k+i}) =& C(A\varphi(x_{k+i|k})+Bu_{k+i}\\~&+H( u_{k+i}\otimes\varphi(x_{k+i|k}))),
\end{array}
\end{equation}
where matrices $A,B,H$ and $C$ are computed via~\eqref{eq:koopman_sol}. 
%
The objective function is chosen here to be
\begin{equation}\label{e12}
\ell(x_k,u_k) = \left\| x_k-x_r\right\|_{Q}^{2} + \left\| u_k\right\|_{R}^{2}\;,
\end{equation}
with $Q \in \mathbb{R}^{n \times n}$ and $R \in \mathbb{R}^{m \times m}$ real symmetric positive-semidefinite matrices designed for the nominal case, and $x_r$ a constant reference defining the desired robot pose.

The developed Koopman-NMPC method is summarized in Algorithm~\ref{alg} and contains two phases. 
The first phase considers offline training to estimate the Koopman-based model by solving~\eqref{eq:koopman_sol}. 
Then, a second online phase follows, where the learned model is leveraged for adapting the NMPC scheme at each time step until the robot converges or collides with an obstacle. 
Similar joint offline-online schemes have been found appropriate for Koopman-based control and estimation of prediction error~\cite{shi2021enhancement}.

\floatstyle{spaceruled}
\restylefloat{algorithm}
\begin{algorithm}[!t]
\caption{Koopman-NMPC}\label{alg}
 \begin{algorithmic}[1]
 \renewcommand{\algorithmicrequire}{\textbf{Offline:}}
 \renewcommand{\algorithmicensure}{\textbf{Online:}}
 \REQUIRE Compute $A,B,H,C$ via solving~\eqref{eq:koopman_sol}
 \ENSURE Choose nominal $Q,R,K$ 
  \FOR {$k = 1,2,\cdots$ }
  \STATE $x_0 = x_k$
  \STATE solve (\ref{eq:optim})
  \STATE $u_k=U_0^*$
  \STATE $x_{k+1}=f_b(x_k,u_k)$
   \ENDFOR
 \end{algorithmic} 
\end{algorithm}

\section{Results and Discussion} \label{6}
We evaluate our method in three distinctive cases. 
The first one is a simulated study that assesses the efficacy of model learning in state estimation and control, and compares the method against a standard NMPC controller. 
Second, we test our method with a realistic digital twin in the Gazebo simulation environment. 
Then, we demonstrate its practical feasibility by testing with physical hardware. 

\subsection{Simulated Study, Analysis, and Comparison with Baseline}

To assess the efficacy of model learning and control, we use a differential-drive model as the nominal model. 
We also consider a perturbed model, whereby the linear and angular velocities include a stochastic perturbation term acting upon the control velocities. 
The governing equations are
\begin{equation}\label{eq:diff_drive}
\left\{\begin{array}{l}
\dot{x}=(v+v_t)\cos\theta +v_s \sin\theta \\
\dot{y}=(v+v_t)\sin\theta -v_s \cos\theta \\
\dot{\theta}=\omega +\omega_s  \\
\end{array}\right.\;,
\end{equation}
where $v$ and $\omega$ are the linear and angular velocities of the robot, $v_t$ and $v_s$ represent the stochastic linear velocity perturbation in the forward direction and normal to the forward direction, respectively, and $\omega_s$ is the perturbation acting on the angular velocity. 

In this work, we assume that the velocity perturbations are proportional to the control velocities given by  
$v_s = \alpha v$, $v_t = \beta v$, $\omega_s = \gamma \omega$. 
Parameters $\alpha$, $\beta$, $\gamma$ are assumed to follow the same exponential distribution and are generated as 
$\alpha \sim \text{Exp}(1/\lambda)$,  
$\beta \sim \text{Exp}(1/\lambda)$, and
$\gamma \sim \text{Exp}(1/\lambda)$, 
where $\lambda$ is the rate.~\footnote{~We note here that this is just one way to include stochasticity into the nominal model. The rationale for the given selection is twofold; large deviations in system velocities are not probable (hence the choice of the exponential distribution), while the additive proportional perturbation can match wheel slip~\cite{ryu2011differential}. An in depth analysis of the different types of perturbations affecting the system and the efficacy to learn a corresponding Koopman-based model and use it for control falls outside the scope of this paper and is a direction of future research enabled by this work.} 
Larger values of $\lambda$ correspond to a higher degree of uncertainty. 
In the simulated tests that follow, we use model~\eqref{eq:diff_drive} to 1) generate training trajectories for learning a Koopman-based model during the offline phase, and 2) propagate the actual state of the simulated system while navigating during the online phase (cf. Alg.~\ref{alg}).

\subsubsection{Model Learning}\label{4.1.1}
To obtain the Koopman-bilinear model we consider here the minimalist dictionary $\varphi(x)=[1,x, cos(\psi), sin(\psi)]$, where $\psi$ is the yaw angle in the robot state.
This is a subset of the more general dictionary based on Hermite polynomials and trigonometric functions proposed in~\cite{shi2021acd}. 
We discretize the scaled dynamics using $4^{th}$-order Runge–Kutta with discretization period $T_s = 0.1$\,s. 
We simulate $1000$ trajectories over $1000$ sampling periods. 
The control input for each trajectory is a random signal uniformly distributed on the unit box $[-1, 1]^2$. 
The initial states of the trajectories are generated randomly with uniform distribution over $[-1m, 1m]^3$. 
The Koopman prediction accuracy is quantified by the root mean squared error, 
$RMSE(\%)=100\sqrt\frac{{\sum_{k=1}^{N}(x_k-\hat{x}_k)^2}}{N}$
. 
The aforementioned training parameter selections yielded an RMSE of $5.55\%$. \footnote{~It is worth noting that this accuracy is inherently dependent on the training data dimension and the sampling time (for instance, RMSE increases to $8.17\%$ for $500$ trajectories, $500$ sampling periods, and $T_s=0.01$\,s).}

\begin{figure}[!h]
\vspace{-3pt}
\centering
\includegraphics[trim={4cm, 1cm, 5.5cm, 2cm},clip,scale=0.148]{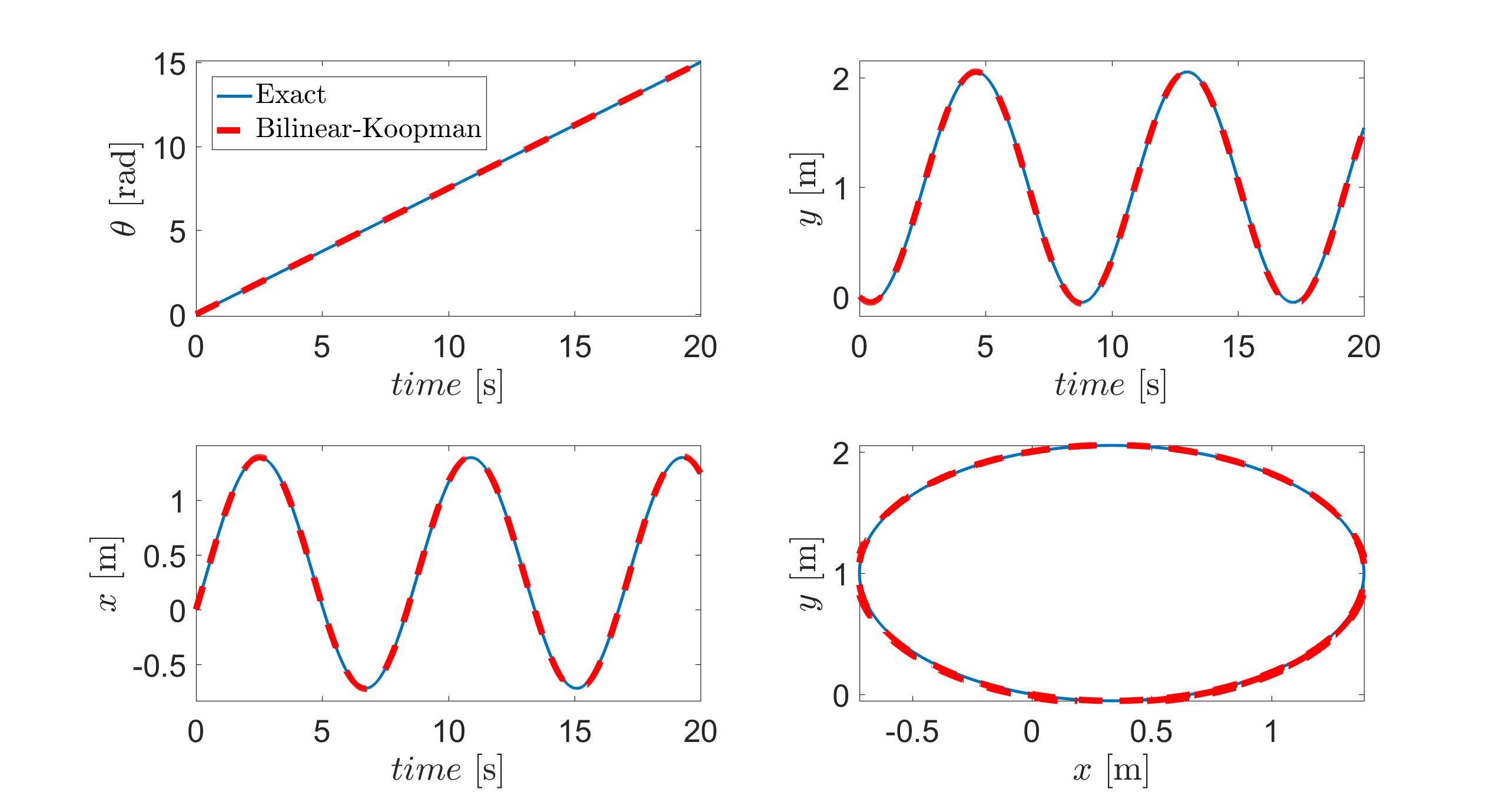}
\vspace{-21pt}
\caption{Comparison of the learned Koopman-bilinear approximation with the exact differential-drive robot model to predict a structured (circular) path. Recall that model learning occurs via randomly-generated control input.}\label{fig1}
\vspace{-6pt}
\end{figure}

To confirm the validity of Koopman-based state estimation we first test with the nominal robot model (i.e. $\alpha=\beta=\gamma=0$),\footnote{~In what follows, we use (the formally undefined) $\lambda=0$ instead to denote the nominal case, for a unified presentation of the different cases.} by setting $v=0.5\,$m/s and $\omega=0.5$\,rad/s. 
Figure~\ref{fig1} confirms that the Koopman-bilinear model can approximate well the exact differential-drive robot model. 

\subsubsection{Koopman-NMPC for the Nominal Robot}
We first test the developed Koopman-NMPC method for the nominal robot case. 
Problem~\eqref{eq:optim} is first posed symbolically using CasADi~\cite{andersson2019casadi}, discretized via multiple-shooting~\cite{leineweber2003efficient}, and solved using IPOPT~\cite{wachter2006implementation}. 
Control and state subspaces $\mathbb{U}$ and $\mathbb{X}$ in~\eqref{eq:optim} represent constraints on decision variables. 
We set actuation limitations to $-0.6\leq v \leq 0.6$, and $-\pi/4 \le \omega \leq  \pi/4$, and consider that the robot is confined in a delimited area of $-2\leq (x,y) \leq 2$. 
%
The NMPC parameters are 
$Q=\begin{bmatrix}
1 & 0 & 0 \\
0 & 5 & 0 \\
0 & 0 & 0.1\\
\end{bmatrix}$, 
$R=\begin{bmatrix}
0.5 & 0  \\
0 & 0.05   \\
\end{bmatrix}$, 
$N=20$, and the update frequency is $10$\,Hz.

\begin{figure}[!t]
\vspace{6pt}
\centering
\includegraphics[trim={0.75cm, 0.35cm, 0.75cm, 0.42cm},clip,scale=0.2165]{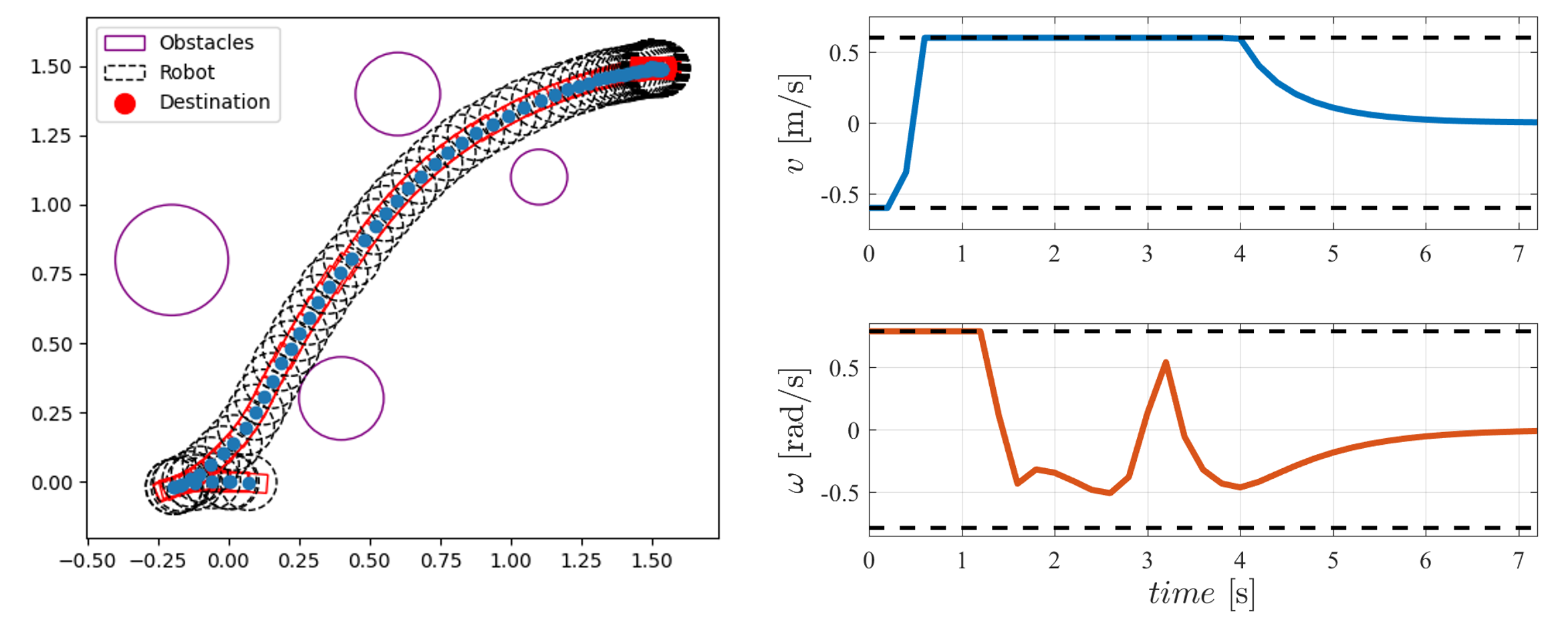}
\vspace{-18pt}
\caption{Resulting trajectory (left) and commanded control velocities (right) for the nominal robot while using the learned Koopman-based model for prediction and control in our Koopman-NMPC method.}\label{fig2}
\vspace{-9pt}
\end{figure}

Figure~\ref{fig2} shows the nominal robot's trajectory and control commands based on the developed Koopman-NMPC method. 
The goal is reached successfully while avoiding collisions. 
We highlight that our method using the Koopman-bilinear learned model matches exactly the optimal trajectory obtained with using directly the nominal nonlinear model (not shown due to lack of space). 
This finding validates the learned model's correctness.

\subsubsection{Standard NMPC for the Perturbed Robot}
We then move on to the control of the robot while being affected by stochastic velocity perturbations.
First, we consider the case of the standard NMPC. 
The nominal (i.e. $\lambda=0$; see footnote 3) model~\eqref{eq:diff_drive} is used by the controller to select commanded controls. 
These are then applied to the perturbed system, that is, model~\eqref{eq:diff_drive} with parameters $\alpha$, $\beta$, and $\gamma$ generated when $\lambda=\{0.01,0.03,0.05,0.07,0.10,0.15,0.20\}$, to propagate the state. 
We perform five trials in each case of rate $\lambda$.

\begin{figure}[!t]
\vspace{6pt}
\captionsetup[subfloat]{farskip=2pt,captionskip=-1pt}
\centering
    \subfloat[]{        \includegraphics[trim={0.80cm, 0.25cm, 1.05cm, 1.25cm},clip,scale=0.38]{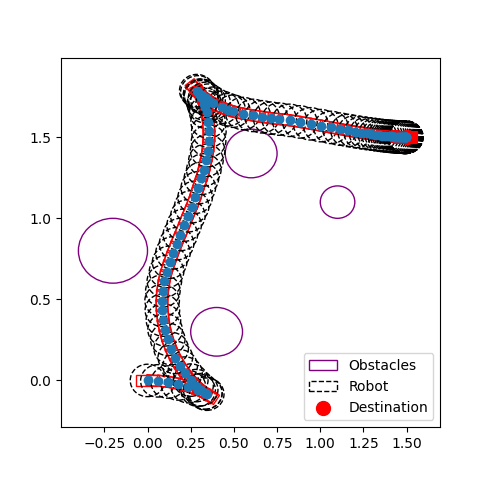}
    }
    \subfloat[]{        \includegraphics[trim={0.88cm, 0.15cm, 1.20cm, 1.25cm},clip,scale=0.335]{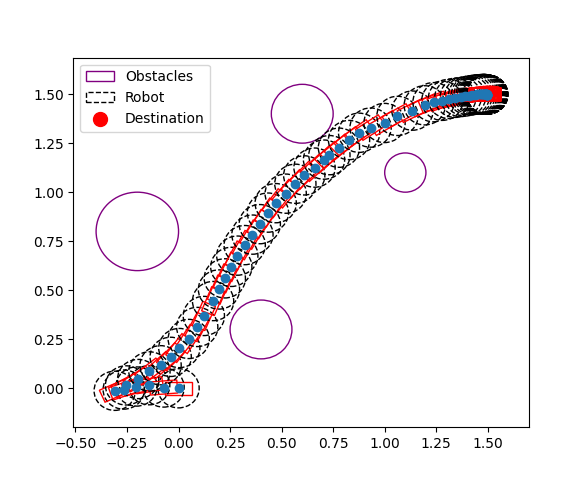}
    }
\vspace{-4pt}
\caption{Resulting trajectories of the robot subject to stochastic velocity perturbations when using (a) nominal NMPC (for $\lambda=0.01$) and (b) our Koopman-NMPC (for $\lambda=0.1$). It can be readily observed that the nominal NMPC is fragile even with very small perturbations (note the collision with the topmost obstacle). In contrast, our method can successfully drive the robot to its goal while avoiding collisions even under larger perturbations.}\label{fig4}
\vspace{-3pt}
\end{figure}

A sample resulting trajectory for the case of $\lambda=0.01$ is visualized in Fig.~\ref{fig4}(a), while detailed results from all conducted trials are shown in Table~\ref{tab1}. 
While the robot is shown to reach its goal, the obstacle avoidance constraint is violated in fact. 
The NMPC problem may thus become infeasible, even under small perturbations affecting the system. 
Increasing the prediction horizon and slightly changing the weights coefficients did not appear to make a difference. 
It is possible to design appropriate terminal costs and constraints to overcome this issue~\cite{grune2017nonlinear}, but it can a challenging task, especially for uncertain systems. 
Instead, enhancing prediction using the Koopman operator is a scalable and more generalizable alternative, as we show next.

Results reported in Table~\ref{tab1} suggest that as system uncertainty increases the success rate decreases (i.e. the collision rate increases). 
The standard NMPC is fragile, with no successful trials observed for values of $\lambda\ge0.05$. 
Among the successful cases, standard NMPC results in trajectories that take longer (on average between $11\%$ for $\lambda=0.01$ to $24\%$ for $\lambda=0.03$) to complete compared to our Koopman-NMPC. 
No significant differences are observed in terms of the average time to collision for both standard NMPC and our Koopman-NMPC, with failed standard NMPC trajectories lasting on average about $10\%$ less compared to Koopman-NMPC failed trials. 
The average time to collision does decrease, as expected, as the system uncertainty increases.

\begin{table}[!t]
\centering
\caption{Koopman-NMPC statistics for simulation.}\label{tab1}
\vspace{-3pt}
\begin{tabular}{c c c c c}
\toprule
\multirow{3}{*}{Method} & \multirow{3}{*}{$\lambda$} & \multirow{3}{0.9cm}{Collision\\ \quad rate} &  Avg.  &  Avg. unsuccessful \\
~ & ~ & ~ & collision-free &  trajectory duration\\
~ & ~ & ~ & traj. duration &  before collision\\
\midrule
NMPC & \multirow{2}{*}{0} &  {0/5} & 8.30s & / \\
K-NMPC & ~ &  {0/5} & 7.30s & / \\
\midrule
NMPC & \multirow{2}{*}{0.01} &  {1/5} & 8.10s & 6.84s \\
K-NMPC & ~ &  {0/5} & 7.30s & / \\
\midrule
NMPC & \multirow{2}{*}{0.03} &  {4/5} & 8.98s & 3.28 \\
K-NMPC & ~ &  {1/5} & 7.23s & 3.20s \\
\midrule
NMPC & \multirow{2}{*}{0.05} &  {5/5} & / & 1.52s \\
K-NMPC & ~ &  {0/5} & 7.05s & / \\
\midrule
NMPC & \multirow{2}{*}{0.07} &  {5/5} & / & 1.52s \\
K-NMPC & ~ &  {1/5} & 6.94s & 1.66s \\
\midrule
NMPC & \multirow{2}{*}{0.1} &  {5/5} & / & 1.50s \\
K-NMPC & ~ &  {1/5} & 6.70s & 1.60s \\
\midrule
NMPC & \multirow{2}{*}{0.15} &  {5/5} & / & 1.40s \\
K-NMPC & ~ &  {3/5} & 6.55s & 1.43s \\
\midrule
NMPC & \multirow{2}{*}{0.2} &  {5/5} & / & 1.36s \\
K-NMPC & ~ &  {1/5} & 6.85s & 1.47s \\
\bottomrule
\end{tabular}
\vspace{-15pt}
\end{table}

\subsubsection{Koopman-NMPC for the Perturbed Robot}


Our developed Koopman-NMPC is found capable to handle successfully cases with uncertainty. 
Figure~\ref{fig4}(b) shows one example with the Koopman-based model trained with data when $\lambda=0.1$ and tested with the same amount of uncertainty. 
Collective results (Table~\ref{tab1}) demonstrate that our method can successfully handle most tested cases, even with higher degrees of system uncertainty. 
Successful collision-free trajectories appear to reach the target faster for higher degrees of uncertainty, likely because the applied stochastic perturbation can yield an additive velocity component on top of the commanded control velocity. 
In all cases depicted in Table~\ref{tab1}, the Koopman-based model is trained with the listed rate $\lambda$, and the computed controls are applied to the perturbed system under the same rate. 

\begin{figure}[!t]
\vspace{6pt}
\centering
  \includegraphics[trim={0.75cm, 0cm, 0.25cm, 0cm},clip,scale=0.222]{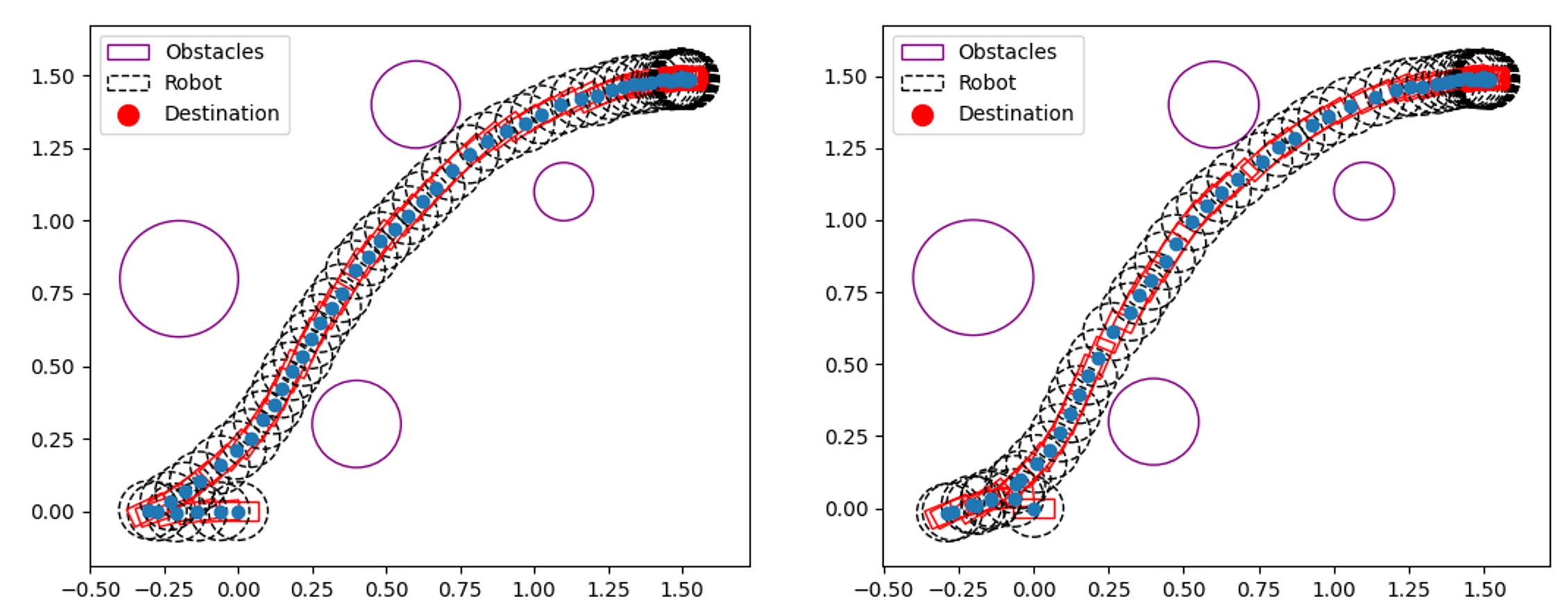}
  \vspace{-12pt}
  \caption{Resulting trajectories of the robot subject to stochastic velocity perturbations when using our developed Koopman-NMPC framework for (left) $\lambda=0.15$ and (right) $\lambda=0.2$. We highlight here that the Koopman-based model used in the depicted cases was trained from data produced with a different rate ($\lambda=0.1$).}\label{fig:kNMPC_gen}
  \vspace{-12pt}
\end{figure}

In an effort to test the ability of our method to generalize, we also present the case of training with $\lambda=0.1$ and using the learned model in deployment under different degrees of uncertainty. 
Figure~\ref{fig:kNMPC_gen} depicts two exemplary cases ($\lambda \in \{0.15,0.20\}$). 
It can be seen that the Koopman-NMPC controller can still drive the robot to its goal. 
The success rates decrease overall in comparison to the case of training and deploying with the same rate $\lambda$ (not shown because of lack of space). 
Overall, however, these are promising findings suggesting the generalization capability of our method; a detailed treatise of this topic as well as a study of the recursive feasibility and stability of the developed Koopman-NMPC technique is part of future work enabled by this work.



\subsection{Testing in Realistic Simulation} \label{7}
Next, we test our Koopman-NMPC method in realistic simulation. 
We use a digital twin of the Husarion ROSbot2.0 Pro robot, a differential-drive wheeled mobile robot. 
(Physical experiments that follow take place with this actual robot.) 
We do not assume knowledge of the system's dynamics, nor any robot-environment interaction variables (e.g., ground friction). 
Instead, a bilinear Koopman model is first learned from data (as outlined earlier), and it is then deployed in the Koopman-NMPC for real-time control. 
All testing runs on a computer with an AMD R5-5600X CPU @ 4.8GHz and 32GB of DRAM @ 3200MHz running the Ubuntu 22.04 operating system. 
All information exchange, control command transmission and overall communications are handled by the Robot Operating System 2 (ROS2) framework~\cite{macenski2022robot}. 

During data collection, the ROSbot model is first loaded in the Gazebo simulator, and then randomized velocity inputs are published via a ROS2 topic to control the robot. 
Corresponding robot poses are published back via another ROS2 topic. 
To prevent the robot from going back and forth frequently, we collect our data in four different scenarios: 1) the linear velocity is non-negative and the angular velocity is non-negative; 2) the linear velocity is non-negative and the angular velocity is non-positive; 3) the linear velocity is non-positive and the angular velocity is non-negative; 4) the linear velocity is non-positive and the angular velocity is non-positive. 
For better integration with ROS2, we set the data collection rate at $10$\,Hz, and each scenario takes $100$\,min, so the total dataset size is 24,000 data points (6,000 for each scenario). 
Velocities are sampled uniformly in the $[0,\pm 1m]^2$ unit square (sign varies according to the specific scenario).

\begin{figure*}[!t]
\vspace{6pt}
\centering
  \includegraphics[scale=0.53]{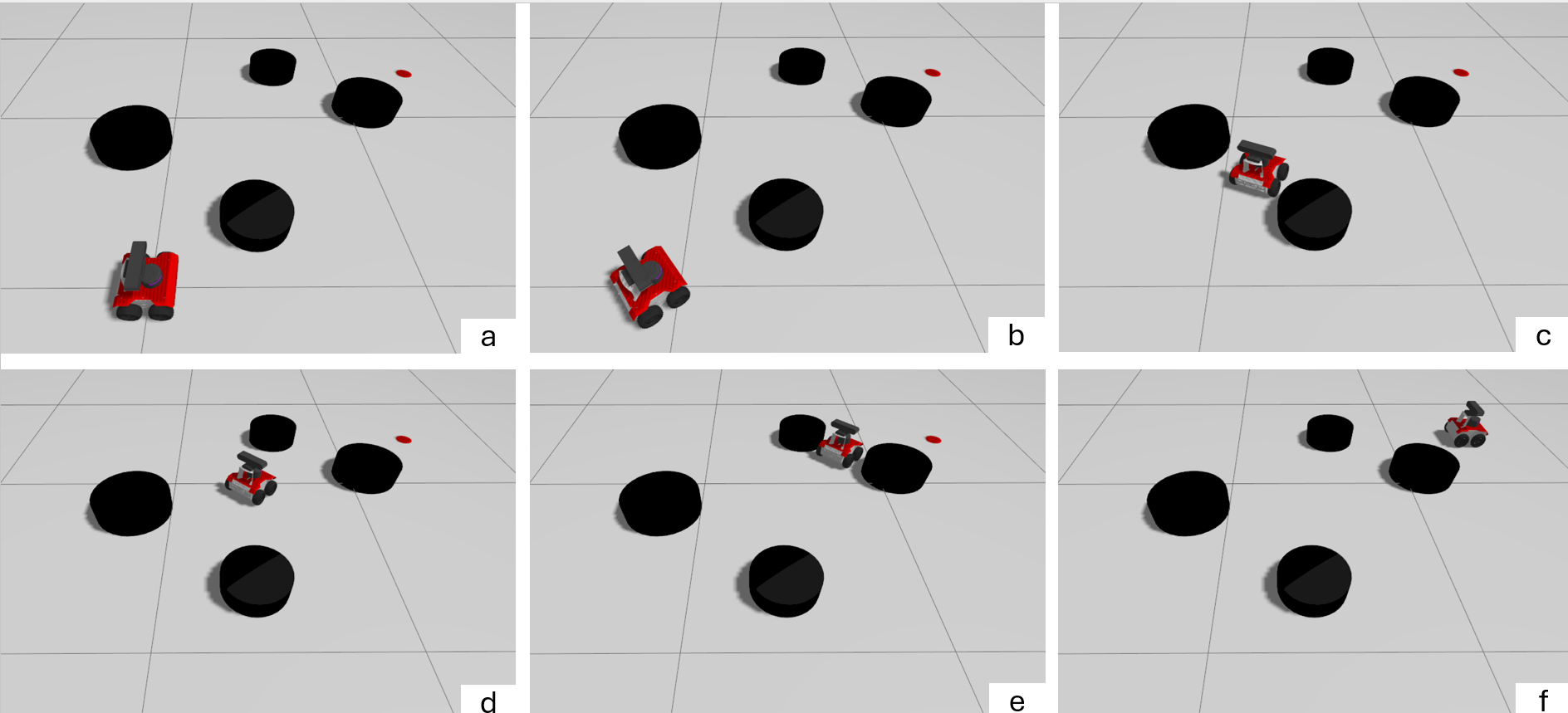}
  \vspace{-6pt}
\caption{Snapshot from one simulated testing case with ROSbot in Gazebo. The goal is depicted with a small circular disk (in red), while obstacles are shown as larger cylinders (in black). Sample trials can be accessed at \protect\url{https://youtu.be/zWRlT7ntFnA}.}\label{fig9}
\vspace{-15pt}
\end{figure*}
\begin{figure}[!t]
\vspace{3pt}
\centering
  \includegraphics[trim={0.40cm, 0cm, 0.20cm, 0cm},clip,scale=0.438]{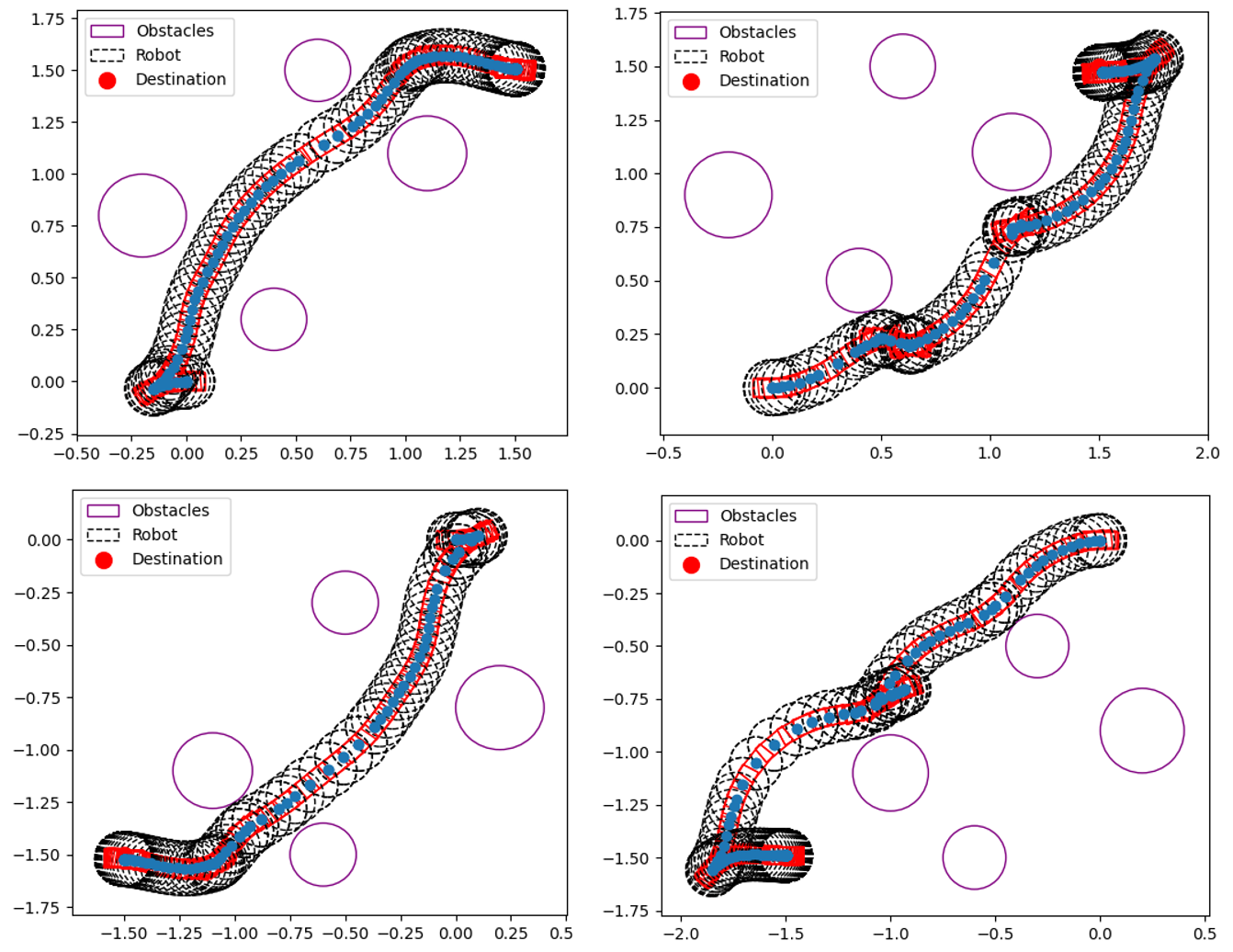}
  \vspace{-18pt}
\caption{Simulated ROSbot trajectories for different obstacle and goal positions. Our method successfully drives the robot to its goal in all cases.}\label{fig10}
\vspace{-12pt}
\end{figure}

Testing takes place on four different maps where the obstacles are arranged differently and the target position varies. 
One of the four cases matches the setup used in the previous set of simulations, and is depicted in Fig.~\ref{fig9} and in the upper left panel of Fig.~\ref{fig10}. 
The former illustration shows snapshots from the simulation as the robot reaches its target, while the latter illustration presents the complete trajectory. 
As before, the robot starts at the origin (0, 0), backs up a little, follows the same trajectory pattern as the virtual robot earlier, successfully passes the obstacles, and finally reaches the goal (1.5, 1.5), with a final position error of around $0.01$\,m. 
The whole process requires $13$\,sec to finish, which is about two times longer than the first set of simulations. 
This can be attributed to unmodeled dynamics (for instance inertia), that is not considered in the differential-drive model~\eqref{eq:diff_drive} and hence not captured during training nor enforced during real-time control in the previous study.


\begin{figure}[!t]
\vspace{6pt}
\centering
  \includegraphics[trim={2cm, 0.5cm, 2.5cm, 1.25cm},clip,scale=0.365]{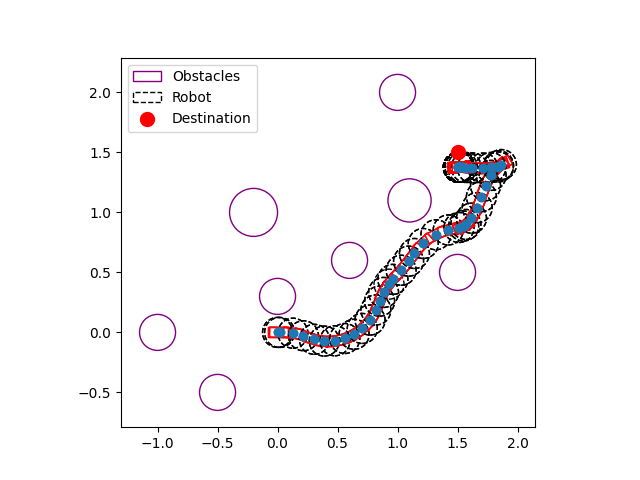}
  \hspace{-6pt}
  \includegraphics[trim={2cm, 0.5cm, 2.5cm, 1.25cm},clip,scale=0.365]{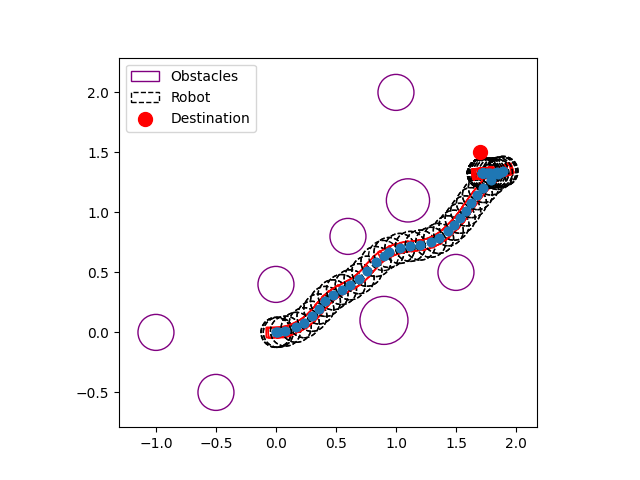}
  \vspace{-21pt}
\caption{Simulated ROSbot trajectories under a more difficult setting.}\label{fig:destination_multiple}
\vspace{-18pt}
\end{figure}

Figure~\ref{fig10} depicts the resulting trajectories in different setups, where the obstacles are arranged differently and the goal position varies to either (1.5, 1.5) or (-1.5, -1.5). 
The initial position remains at (0, 0). 
The initial and goal orientation is set to 0 as well, in all cases. 
Our method is observed to drive the system to its goal while avoiding collisions in all cases. 
We further test the proposed method under two more complex maps, where the starting and desired poses remain the same, but the number of obstacles is doubled, as shown in Fig.~\ref{fig:destination_multiple}. 
In this case, the success rates can vary depending on obstacle proximity and arrangement. 
In the first case (left panel of the figure), all five conducted tests were successful, whereas in the second case (right panel), the success rate was $60\%$ (three collision-free trials out of five). 
The main difference between the two maps is that the robot has to go through multiple objects very close to each other in the second map, thus necessitating small but frequent heading variations. 
We hypothesize that such heading variations are hindering the ability of the system to adapt fast enough, hence leading to collisions (as in the second case) and increased final position error (the average total position error in these experiments is $0.09$\;m). 
Improving the performance under more complex environments is part of future work. 

\subsection{Validation via Physical Hardware Experiments}
Lastly yet importantly, we validate our method with physical hardware experiments.
We use the physical ROSbot2.0 Pro robot, and the same computer as in the simulation testing. 
A motion capture camera system is used to track the pose of the robot and provide this information to the controller in real time. 
The obstacles are also tracked in motion capture, in an effort to streamline data collection. 

\begin{figure*}[!t]
\vspace{6pt}
\captionsetup[subfigure]{labelformat=empty}
\centering
\subfloat[]{
\includegraphics[trim={0cm, 10cm, 10cm, 22cm},clip,width=0.188\textwidth]{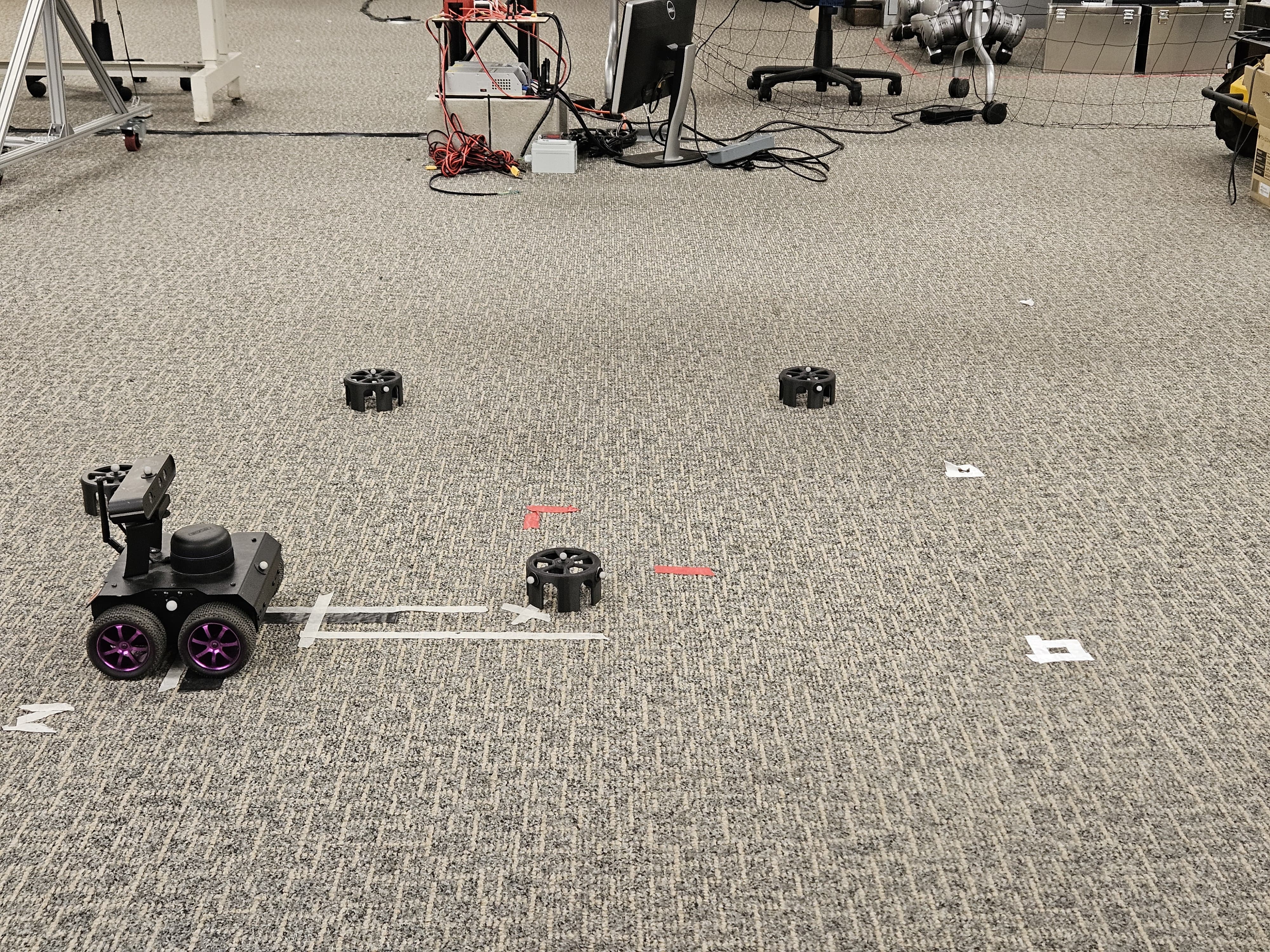}
}
\subfloat[]{
\includegraphics[trim={0cm, 10cm, 10cm, 22cm},clip,width=0.188\textwidth]{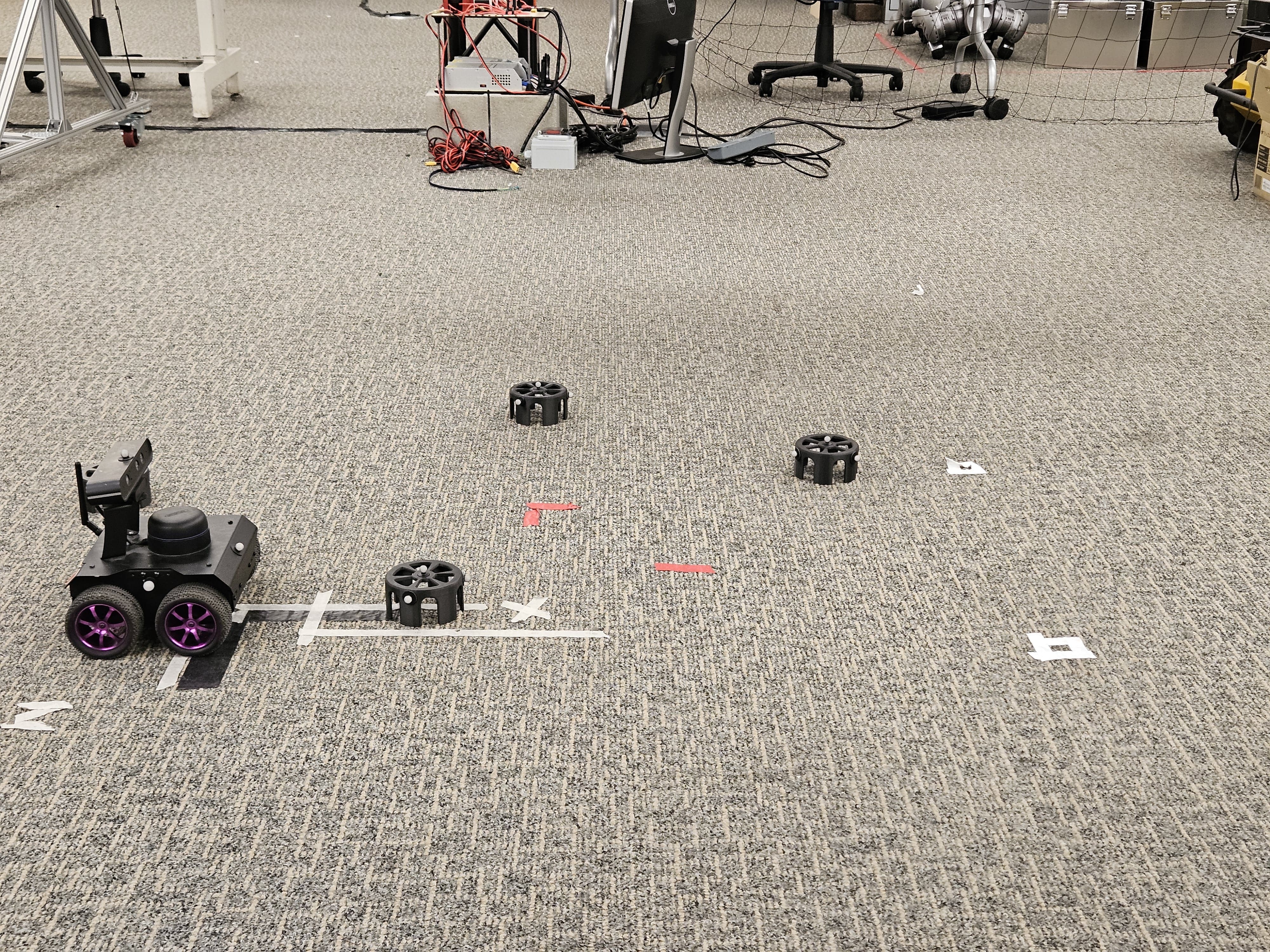}
}
\subfloat[]{
\includegraphics[trim={0cm, 10cm, 10cm, 22cm},clip,width=0.188\textwidth]{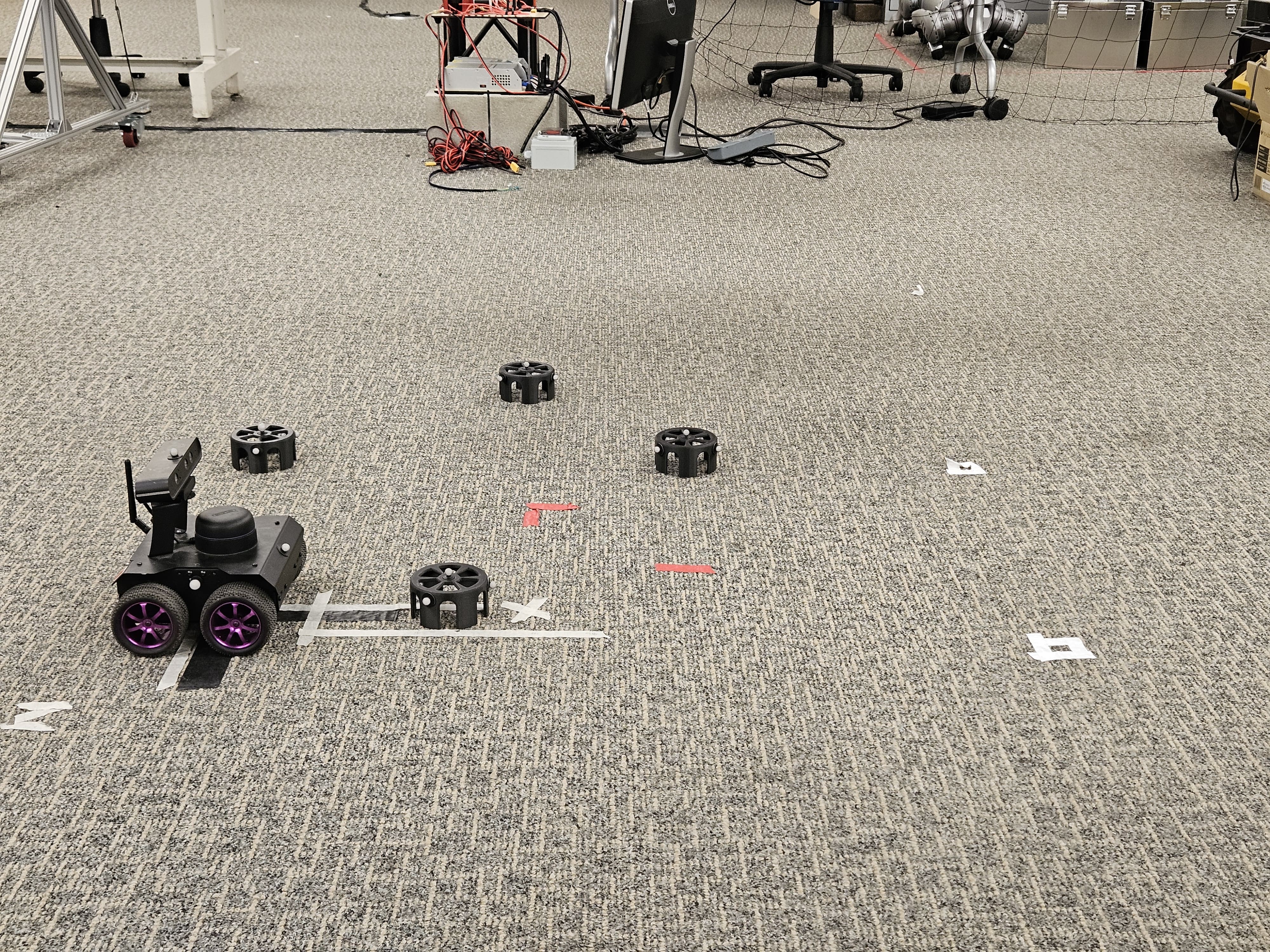}
}
\subfloat[]{
\includegraphics[trim={0cm, 10cm, 10cm, 22cm},clip,width=0.188\textwidth]{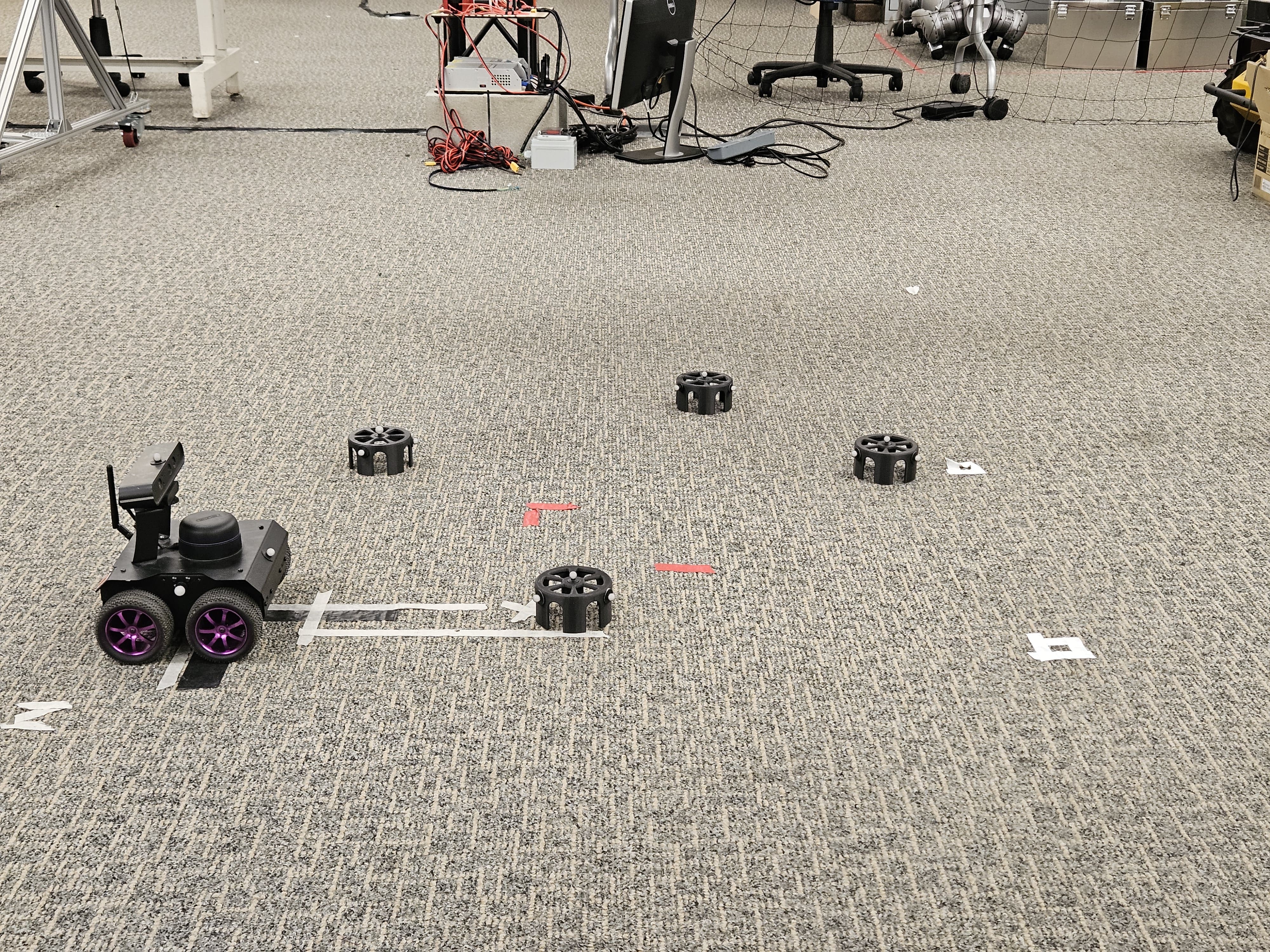}
}
\subfloat[]{
\includegraphics[trim={0cm, 10cm, 10cm, 22cm},clip,width=0.188\textwidth]{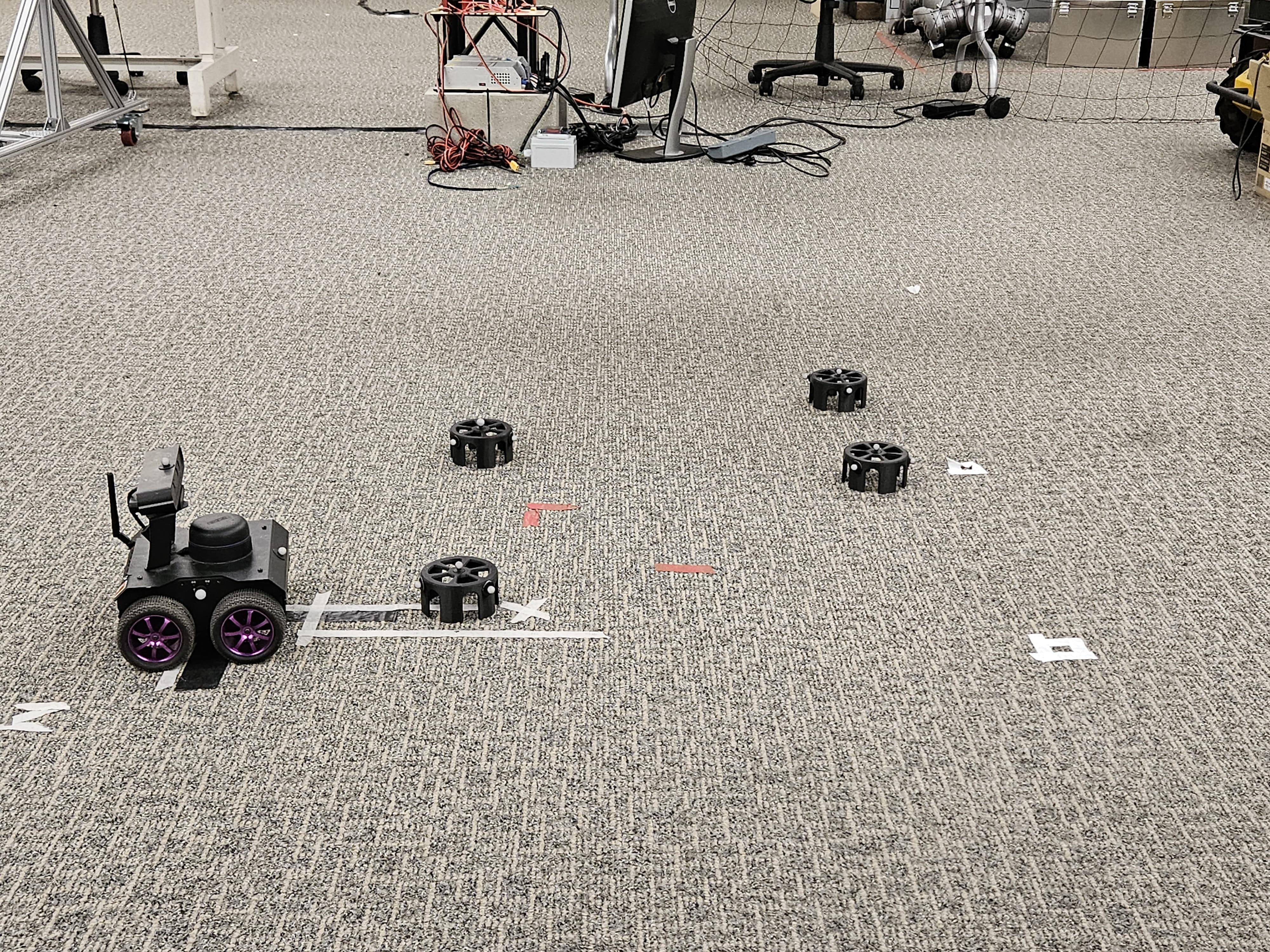}
}
\\
\vspace{-18pt}
\subfloat[]{
\includegraphics[trim={0.5cm, 0.2cm, 1.0cm, 1cm},clip,width=0.188\textwidth]{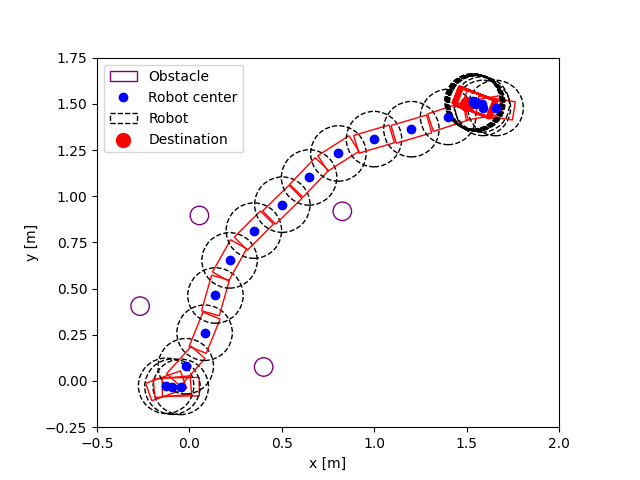}
}
\subfloat[]{
\includegraphics[trim={0.5cm, 0.2cm, 1.0cm, 1cm},clip,width=0.188\textwidth]{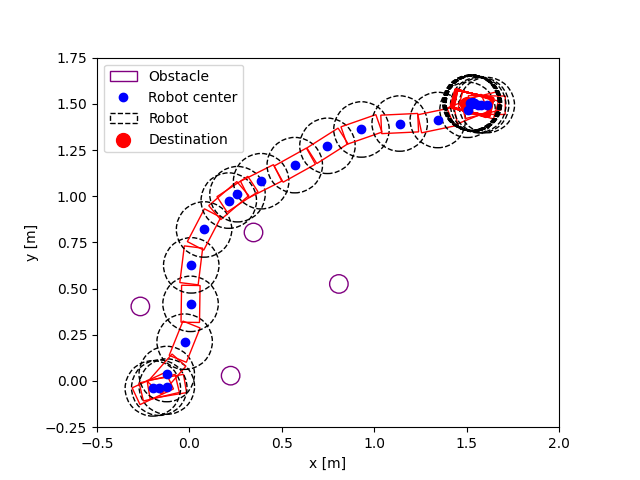}
}
\subfloat[]{
\includegraphics[trim={0.5cm, 0.2cm, 1.0cm, 1cm},clip,width=0.188\textwidth]{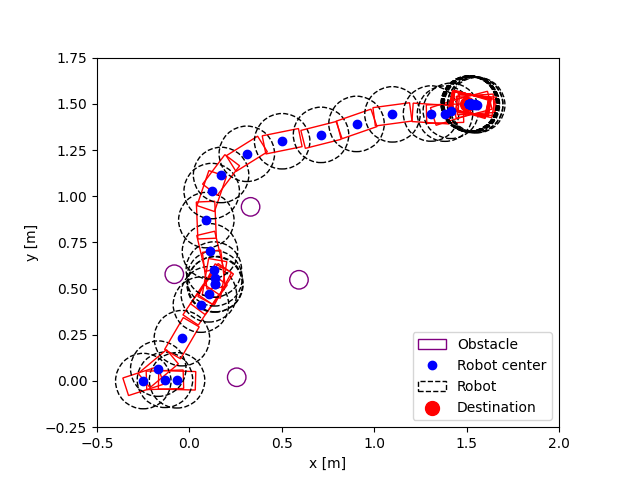}
}
\subfloat[]{
\includegraphics[trim={0.5cm, 0.2cm, 1.0cm, 1cm},clip,width=0.188\textwidth]{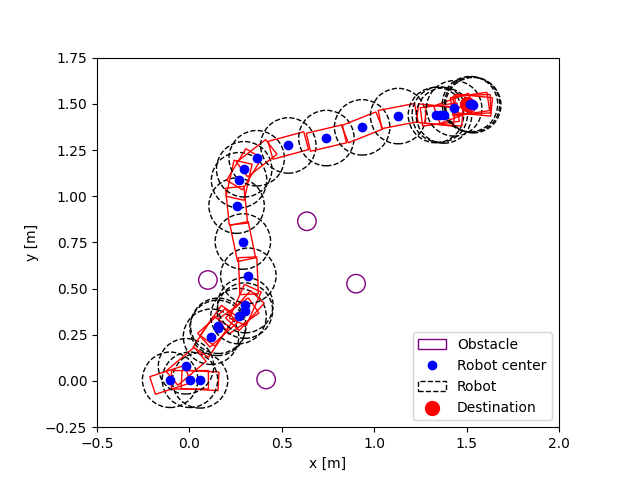}
}
\subfloat[]{
\includegraphics[trim={0.5cm, 0.2cm, 1.0cm, 1cm},clip,width=0.188\textwidth]{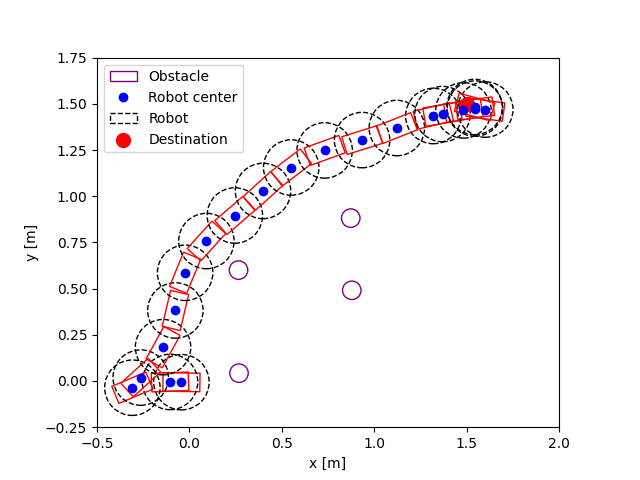}
}
\\
\vspace{0pt}
\subfloat[]{
\includegraphics[trim={0cm, 10cm, 10cm, 22cm},clip,width=0.188\textwidth]{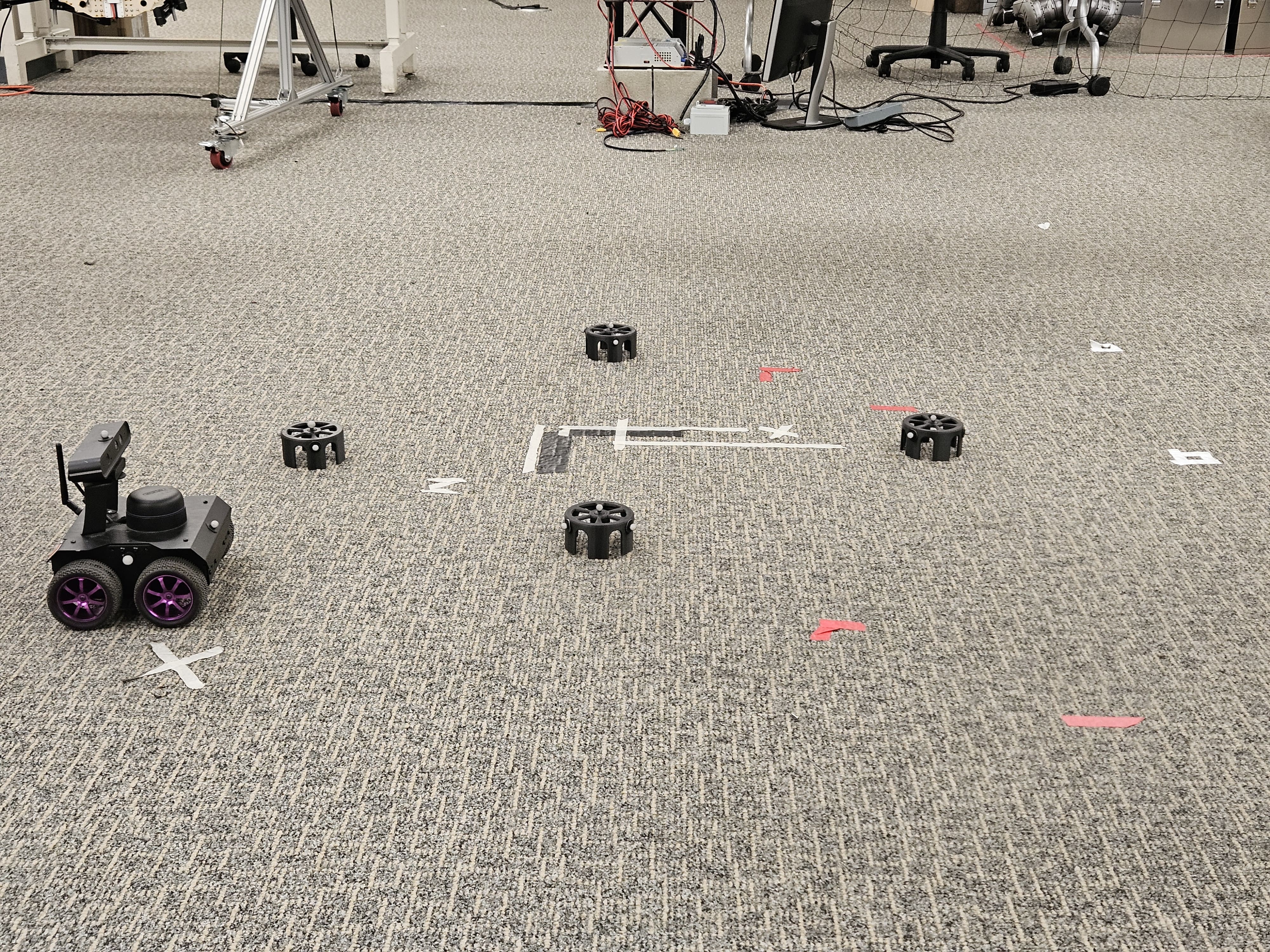}
}
\subfloat[]{
\includegraphics[trim={0cm, 10cm, 10cm, 22cm},clip,width=0.188\textwidth]{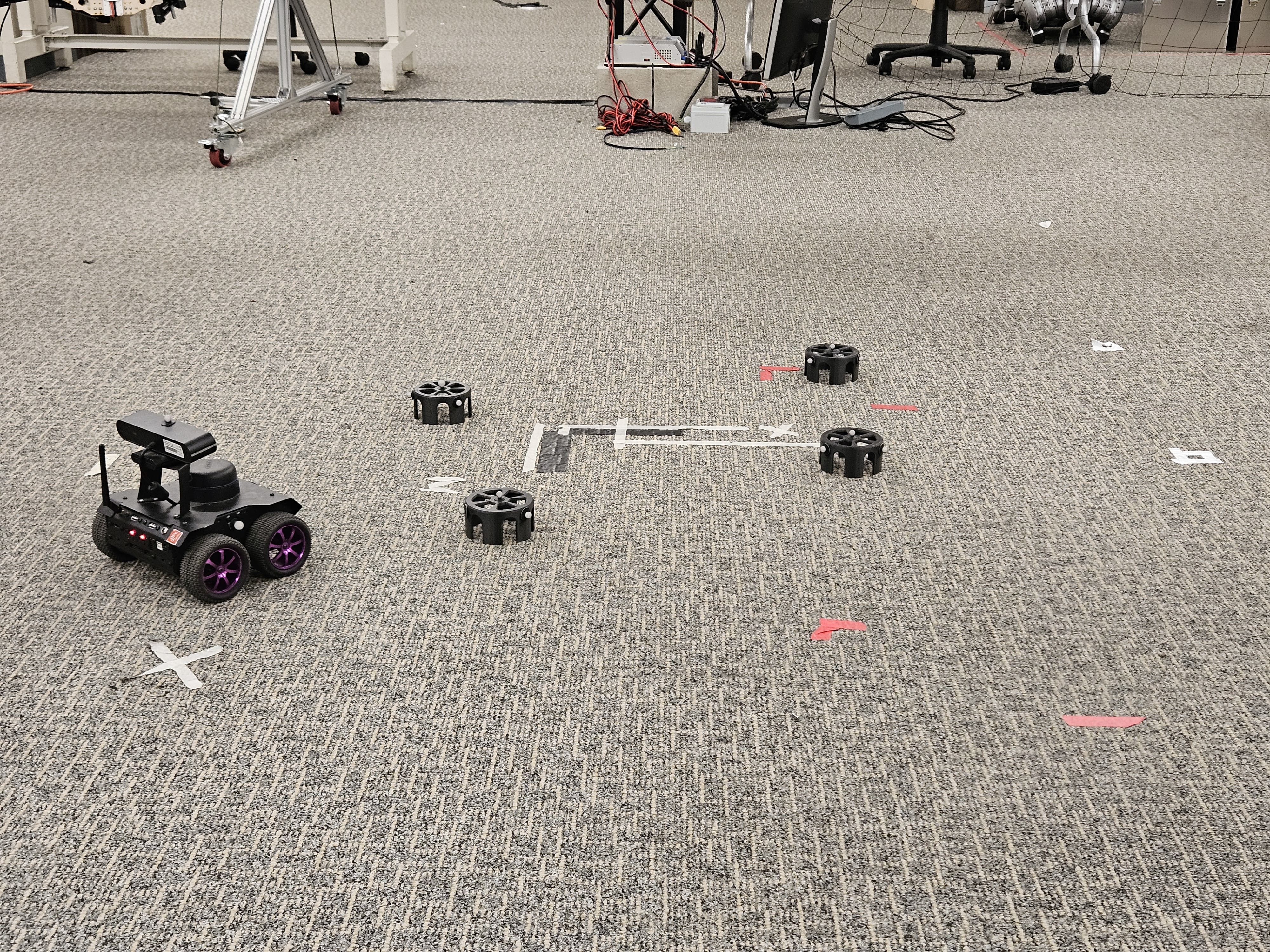}
}
\subfloat[]{
\includegraphics[trim={0cm, 10cm, 10cm, 22cm},clip,width=0.188\textwidth]{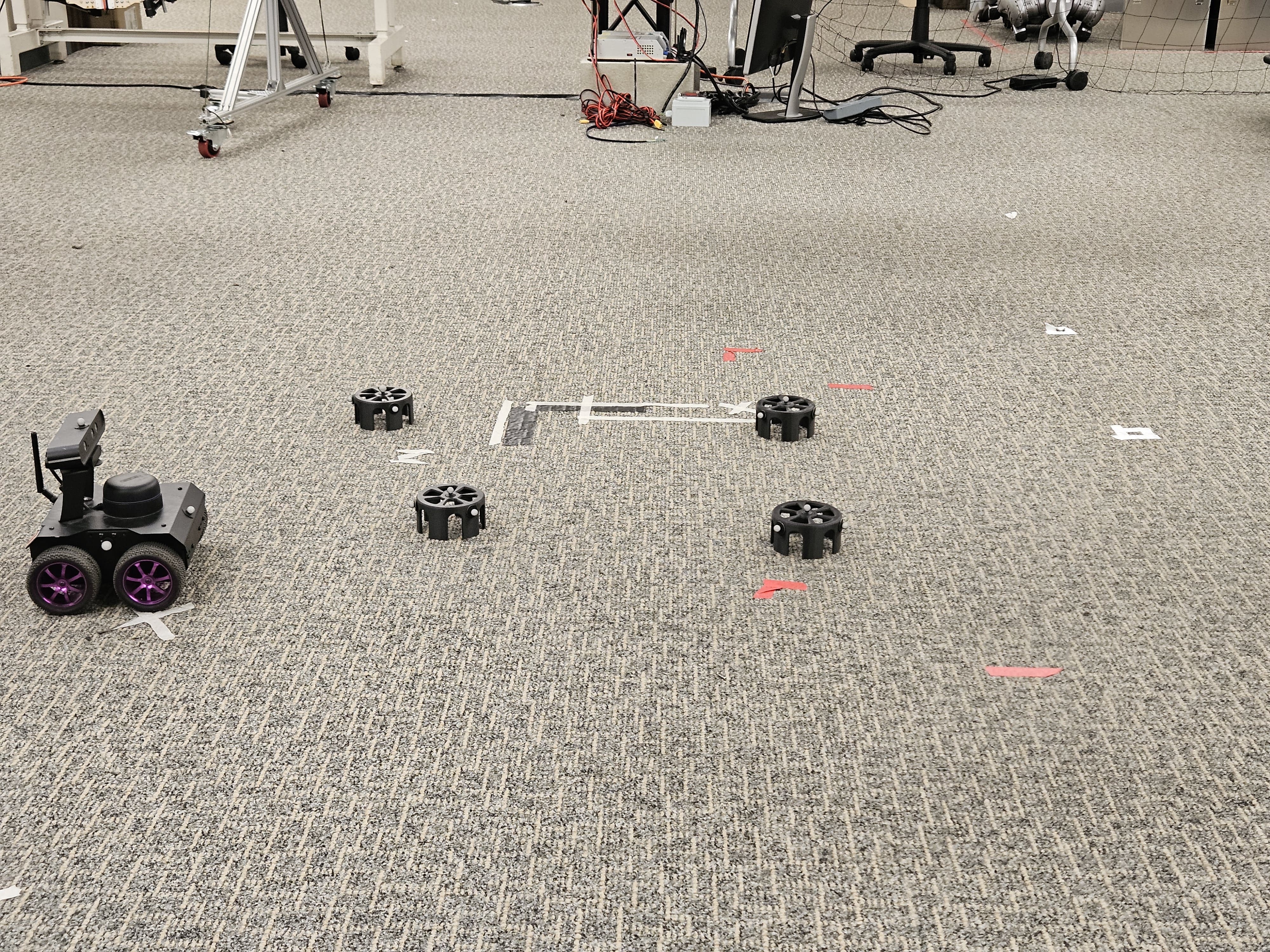}
}
\subfloat[]{
\includegraphics[trim={0cm, 10cm, 10cm, 22cm},clip,width=0.188\textwidth]{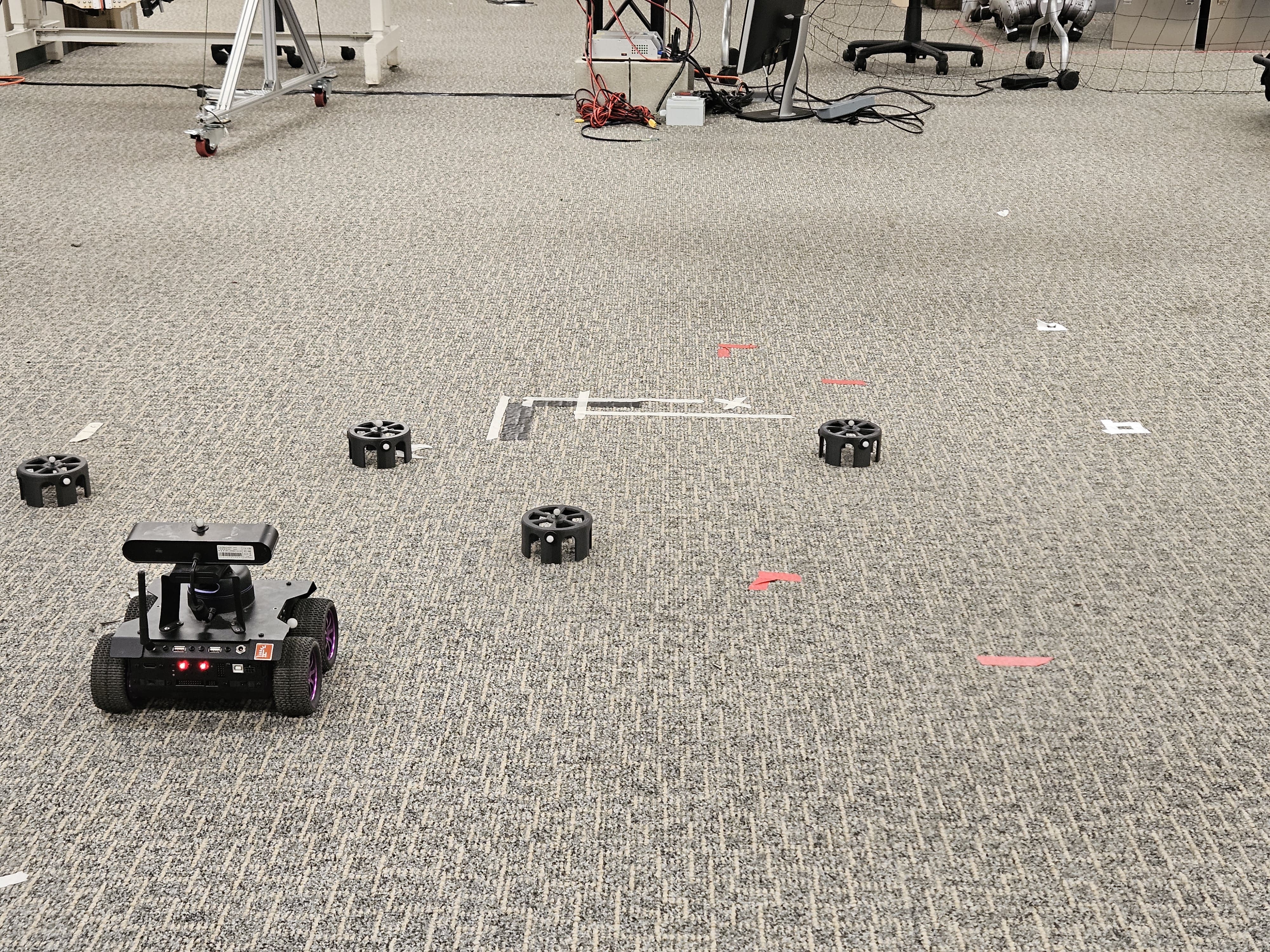}
}
\subfloat[]{
\includegraphics[trim={0cm, 10cm, 10cm, 22cm},clip,width=0.188\textwidth]{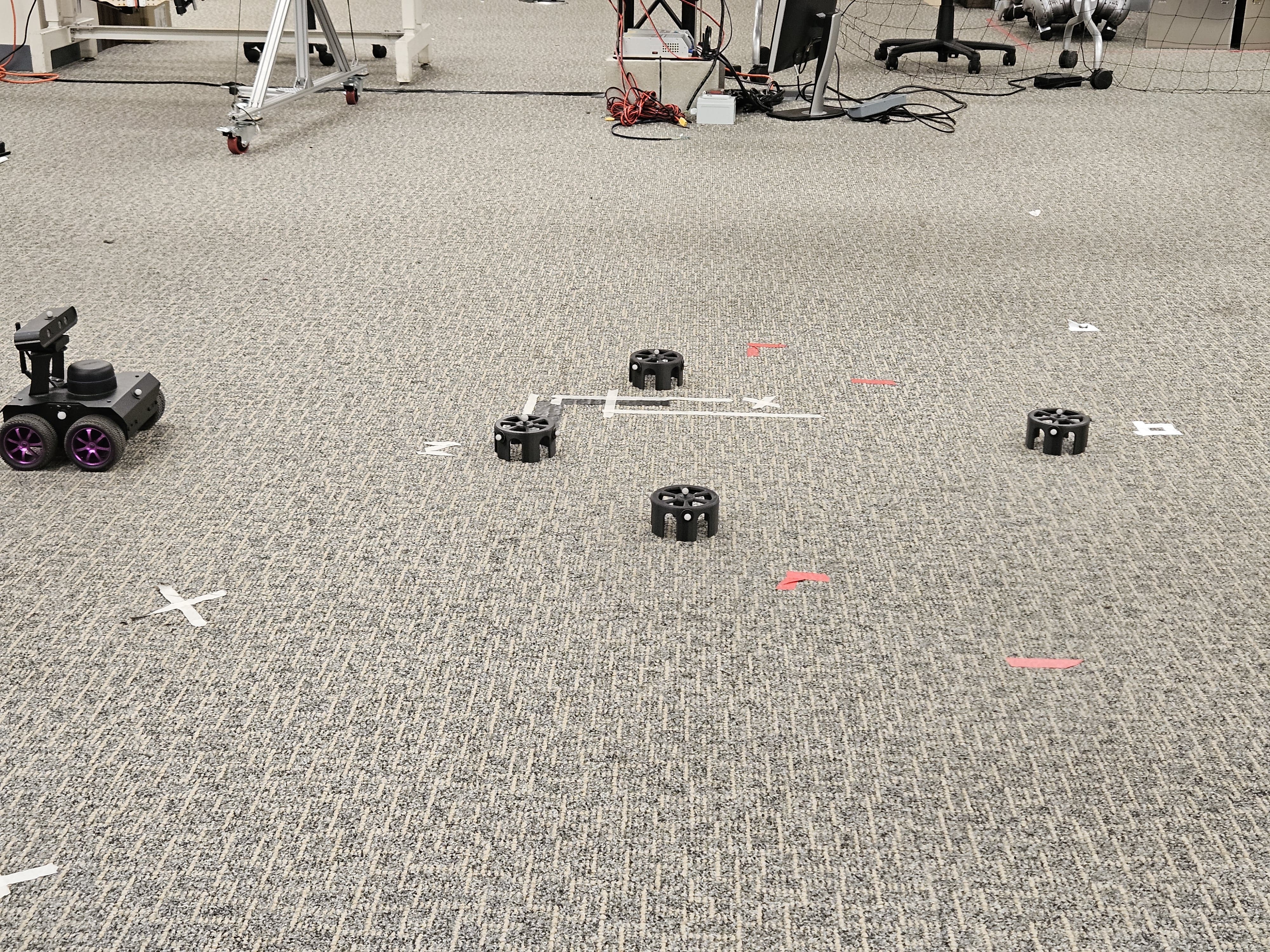}
}
\\
\vspace{-18pt}
\subfloat[]{
\includegraphics[trim={0.5cm, 0.2cm, 1.2cm, 1.2cm},clip,width=0.188\textwidth]{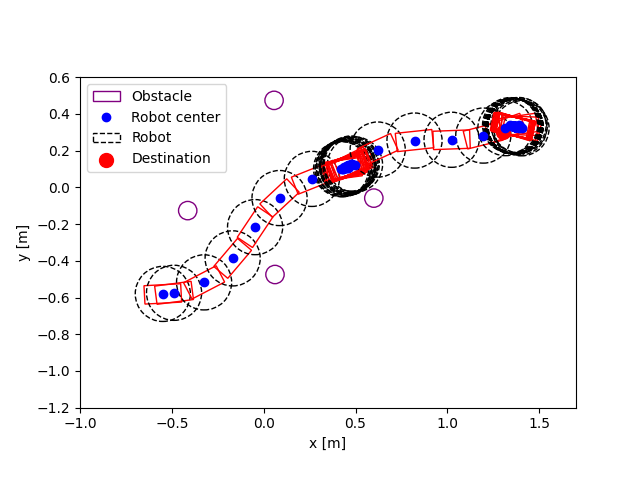}
}
\subfloat[]{
\includegraphics[trim={0.5cm, 0.2cm, 1.2cm, 1.2cm},clip,width=0.188\textwidth]{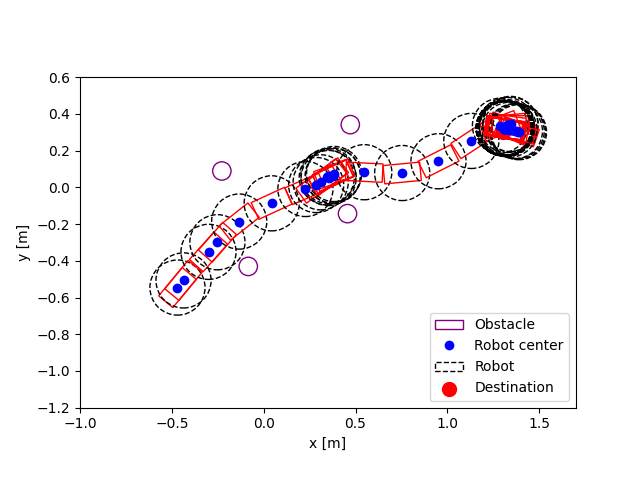}
}
\subfloat[]{
\includegraphics[trim={0.5cm, 0.2cm, 1.2cm, 1.2cm},clip,width=0.188\textwidth]{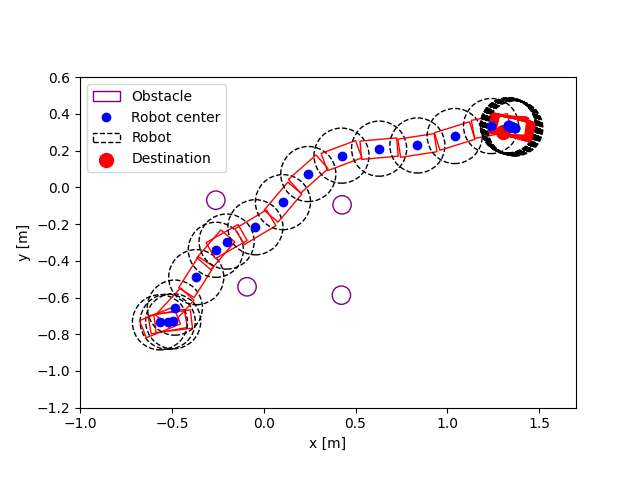}
}
\subfloat[]{
\includegraphics[trim={0.5cm, 0.2cm, 1.2cm, 1.2cm},clip,width=0.188\textwidth]{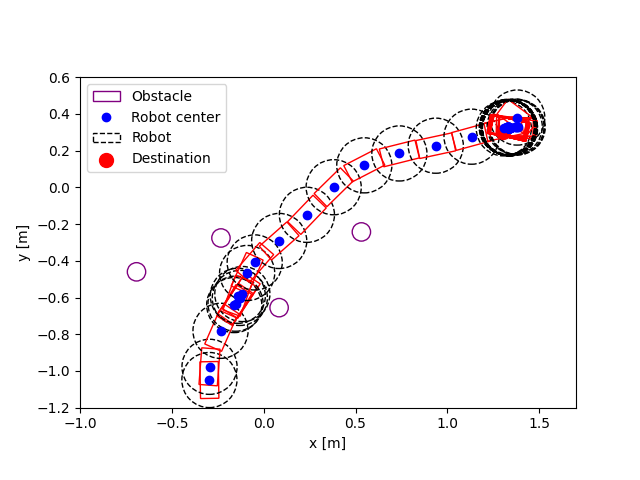}
}
\subfloat[]{
\includegraphics[trim={0.5cm, 0.2cm, 1.2cm, 1.2cm},clip,width=0.188\textwidth]{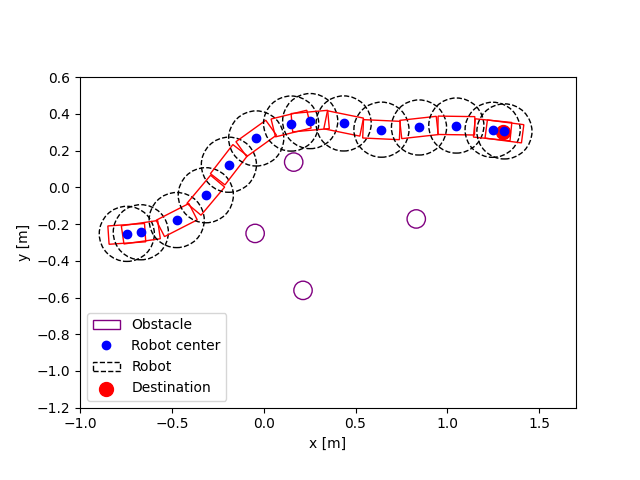}
}
\vspace{-15pt}
\caption{Sample physical experiments over 10 different maps (obstacle arrangement) and varying initial pose (in the cases depicted in the third row). The first and third rows depict the initialization, while rows two and four show one successful trajectory in the corresponding map directly above each panel.}\label{fig:physical_experiments}
\vspace{-14pt}
\end{figure*}

The operating environment consists of four obstacles but smaller than those in simulation, to adhere to the available physical space boundaries. 
A total of 10 different obstacle arrangements are considered, shown in Fig.~\ref{fig:physical_experiments}. 
In each map we conduct five trials. 
The initial and goal poses remain the same across the trials in each map. 
In the first five cases (first row), the initial and goal poses are the same in all maps.
In the latter five cases (third row), the initial pose varies with each map but the goal remains the same. 
These settings give a rich set of conditions used to validate our method.



To address latencies within the whole system, the control frequency of the robot is lowered to $1$\,Hz for both data collection and experiments. 
To avoid frequent abrupt velocity changes of the robot, in data collection the random velocity commands follow a normal distribution $V,\Omega \sim \mathcal{N}(0,0.1^2)$. 
The Koopman operator is trained with a dataset of size 6200 (103 min of data). 
As in Gazebo simulations, we do not assume knowledge of the system's dynamics, nor any robot-environment interaction variables (e.g., ground friction). 
Further, control signal and networking latencies, as well as terrain variations not present in simulation, are additional sources of uncertainty that affect model learning and real-time control. 
Sample trials can be accessed at \url{https://youtu.be/zWRlT7ntFnA}.

Results are shown in Fig.~\ref{fig:physical_experiments} and Table~\ref{tab2}. 
The 10 different maps as well as the initial pose in each case are shown, along with a sample trajectory of the robot successfully reaching the given goal while avoiding obstacles, using the developed Koopman-NMPC method. 
Collective results reveal that our method can drive the robot to its goal with high ($88\%$) success rates overall; in six out of 10 maps no collisions occurred. 
The average time to goal remains similar for the majority of the tested cases. 
Map 3 (middle column of first and second rows in Fig.~\ref{fig:physical_experiments}) is the only exception, likely because of the tight arrangement of obstacles early in the trajectory (a higher collision rate was also observed in this scenario). 
The time it takes the robot to reach the destination is reflected by the number of circles in the plots (the update rate is fixed at $1$\;Hz). 
Experiments last a minimum of $21$\;sec and a maximum of $60$\;sec. 
These findings support the practical utility of the Koopman-NMPC to quickly and effectively learn a previously unknown dynamics model to be used in wheeled mobile robot closed-loop navigation control. 

\begin{table}[!t]
\vspace{6pt}
\centering
\caption{Koopman-NMPC statistics for physical experiments.}\label{tab2}
\vspace{-3pt}
\begin{tabular}{c c c c}
\toprule
\multirow{3}{*}{Map} & \multirow{3}{*}{Collision rate} &  Avg.  &  Avg. unsuccessful \\
 ~ & ~ & collision-free &  trajectory duration\\
 ~ & ~ & traj. duration &  before collision\\
\midrule
1 &  {0/5} & 30.1s & / \\
2 &  {1/5} & 35.5s & 7.8s \\
3 &  {2/5} & 52.2s & 6.3s \\
4 &  {0/5} & 28.9s & / \\
5 &  {2/5} & 28.7s & 5.3s \\
6 &  {0/5} & 25.4s & / \\
7 &  {0/5} & 23.2s & / \\
8 &  {0/5} & 25.1s & / \\
9 &  {0/5} & 26.6s & / \\
10 &  {1/5} & 26.1s & 7.1s \\
\bottomrule
\end{tabular}
\vspace{-12pt}
\end{table}

\section{Conclusion} \label{8}
In this paper, a Koopman-based NMPC framework was developed and applied for closed-loop navigation control of a wheeled mobile robot under the presence of uncertainty. 
A Koopman-bilinear model served as the learned representation of an underlying stochastic control-affine system (differential drive with additive stochastic velocity perturbations). 
A formulation based on the Kronecker product enabled the computationally-efficient estimation of the Koopman-bilinear system by solving a dual least-squares problem based on training data collected offline. 
The obtained Koopman-bilinear model was shown capable to handle system uncertainty well and fit into real-time NMPC design to drive the perturbed robot to a desired target while avoiding obstacles.  
The formulated NMPC allows for fewer decision variables and a meaningful design of the state weight matrix. 
Simulation results show the effect of our method over standard NMPC, which is fragile to the considered type of uncertainty. 
Realistic digital twin testing and physical hardware experiments demonstrate the utility and practical feasibility of our method for learning-based mobile robot navigation.

This work presents an example of model learning for mobile robot control to handle uncertainty during deployment, and demonstrates how an offline training phase can be used to estimate a low-dimensional model that can serve in control design. 
Several directions of future work enabled by this paper are possible. 
Examples include the development and deployment into robots with higher dimensional dynamics (e.g., aerial robots), inclusion of different types of uncertainty, study of the stability, recursive feasibility and scalability of the proposed method, and analysis of the effect of learned model errors into control design. 
Further, a study of the sim-to-real behavior of the learned model can also pave the way for online and active learning of model uncertainty.



\bibliographystyle{IEEEtran} 
\bibliography{ref}

\begin{thebibliography}{10}
\providecommand{\url}[1]{#1}
\csname url@samestyle\endcsname
\providecommand{\newblock}{\relax}
\providecommand{\bibinfo}[2]{#2}
\providecommand{\BIBentrySTDinterwordspacing}{\spaceskip=0pt\relax}
\providecommand{\BIBentryALTinterwordstretchfactor}{4}
\providecommand{\BIBentryALTinterwordspacing}{\spaceskip=\fontdimen2\font plus
\BIBentryALTinterwordstretchfactor\fontdimen3\font minus \fontdimen4\font\relax}
\providecommand{\BIBforeignlanguage}[2]{{%
\expandafter\ifx\csname l@#1\endcsname\relax
\typeout{** WARNING: IEEEtran.bst: No hyphenation pattern has been}%
\typeout{** loaded for the language `#1'. Using the pattern for}%
\typeout{** the default language instead.}%
\else
\language=\csname l@#1\endcsname
\fi
#2}}
\providecommand{\BIBdecl}{\relax}
\BIBdecl

\bibitem{wang2008modeling}
D.~Wang and C.~B. Low, ``Modeling and analysis of skidding and slipping in wheeled mobile robots: Control design perspective,'' \emph{IEEE Transactions on Robotics}, vol.~24, no.~3, pp. 676--687, 2008.

\bibitem{ji2022proactive}
T.~Ji, A.~N. Sivakumar, G.~Chowdhary, and K.~Driggs-Campbell, ``Proactive anomaly detection for robot navigation with multi-sensor fusion,'' \emph{IEEE Robotics and Automation Letters}, vol.~7, no.~2, pp. 4975--4982, 2022.

\bibitem{li2015trajectoryMPC}
Z.~Li, J.~Deng, R.~Lu, Y.~Xu, J.~Bai, and C.-Y. Su, ``Trajectory-tracking control of mobile robot systems incorporating neural-dynamic optimized model predictive approach,'' \emph{Transactions on Systems, Man, and Cybernetics: Systems}, vol.~46, no.~6, pp. 740--749, 2015.

\bibitem{pacheco2015MPC}
L.~Pacheco and N.~Luo, ``Testing pid and mpc performance for mobile robot local path-following,'' \emph{International Journal of Advanced Robotic Systems}, vol.~12, no.~11, p. 155, 2015.

\bibitem{grune2017nonlinear}
L.~Gr{\"u}ne, J.~Pannek, L.~Gr{\"u}ne, and J.~Pannek, \emph{Nonlinear model predictive control}.\hskip 1em plus 0.5em minus 0.4em\relax Springer, 2017.

\bibitem{fleming2014robust}
J.~Fleming, B.~Kouvaritakis, and M.~Cannon, ``Robust tube mpc for linear systems with multiplicative uncertainty,'' \emph{IEEE Transactions on Automatic Control}, vol.~60, no.~4, pp. 1087--1092, 2014.

\bibitem{limon2006input}
D.~Lim{\'o}n, T.~Alamo, F.~Salas, and E.~F. Camacho, ``Input to state stability of min--max mpc controllers for nonlinear systems with bounded uncertainties,'' \emph{Automatica}, vol.~42, no.~5, pp. 797--803, 2006.

\bibitem{sun2019self}
Z.~Sun, V.~Rostampour, and M.~Cao, ``Self-triggered stochastic mpc for linear systems with disturbances,'' \emph{IEEE Control Systems Letters}, vol.~3, no.~4, pp. 787--792, 2019.

\bibitem{edwards2021automatic}
W.~Edwards, G.~Tang, G.~Mamakoukas, T.~Murphey, and K.~Hauser, ``Automatic tuning for data-driven model predictive control,'' in \emph{IEEE International Conference on Robotics and Automation (ICRA)}, 2021, pp. 7379--7385.

\bibitem{karydis2015probabilistically}
K.~Karydis, I.~Poulakakis, J.~Sun, and H.~G. Tanner, ``Probabilistically valid stochastic extensions of deterministic models for systems with uncertainty,'' \emph{The International Journal of Robotics Research}, vol.~34, no.~10, pp. 1278--1295, 2015.

\bibitem{karoly2020deep}
A.~I. K{\'a}roly, P.~Galambos, J.~Kuti, and I.~J. Rudas, ``Deep learning in robotics: Survey on model structures and training strategies,'' \emph{IEEE Transactions on Systems, Man, and Cybernetics: Systems}, vol.~51, no.~1, pp. 266--279, 2020.

\bibitem{oikawa2021reinforcement}
M.~Oikawa, T.~Kusakabe, K.~Kutsuzawa, S.~Sakaino, and T.~Tsuji, ``Reinforcement learning for robotic assembly using non-diagonal stiffness matrix,'' \emph{IEEE Robotics and Automation Letters}, vol.~6, no.~2, pp. 2737--2744, 2021.

\bibitem{chen2024llm}
W.~Chen, G.~Li, M.~Li, W.~Wang, P.~Li, X.~Xue, X.~Zhao, and L.~Liu, ``Llm-enabled incremental learning framework for hand exoskeleton control,'' \emph{IEEE Transactions on Automation Science and Engineering}, 2024.

\bibitem{zhao2024deep}
D.~Zhao, B.~Li, F.~Lu, J.~She, and S.~Yan, ``Deep bilinear koopman model predictive control for nonlinear dynamical systems,'' \emph{IEEE Transactions on Industrial Electronics}, 2024.

\bibitem{shi2024koopman}
L.~Shi, M.~Haseli, G.~Mamakoukas, D.~Bruder, I.~Abraham, T.~Murphey, J.~Cortes, and K.~Karydis, ``Koopman operators in robot learning,'' \emph{arXiv preprint arXiv:2408.04200}, 2024.

\bibitem{zhou2025learning}
J.~Zhou, Y.~Zhu, X.~Zhang, S.~Agrawal, and K.~Karydis, ``Learning-based estimation of forward kinematics for an orthotic parallel robotic mechanism,'' in \emph{19th IEEE/RAS-EMBS International Conference on Rehabilitation Robotics (ICORR)}, 2025, pp. 1--6.

\bibitem{korda2018linear}
M.~Korda and I.~Mezi{\'c}, ``Linear predictors for nonlinear dynamical systems: Koopman operator meets model predictive control,'' \emph{Automatica}, vol.~93, pp. 149--160, 2018.

\bibitem{shi2023koopman}
L.~Shi, Z.~Liu, and K.~Karydis, ``Koopman operators for modeling and control of soft robotics,'' \emph{Current Robotics Reports}, vol.~4, no.~2, pp. 23--31, 2023.

\bibitem{mamakoukas2021Taylor}
G.~Mamakoukas, M.~L. Castano, X.~Tan, and T.~D. Murphey, ``Derivative-based koopman operators for real-time control of robotic systems,'' \emph{IEEE Transactions on Robotics}, 2021.

\bibitem{shi2020data}
L.~Shi, H.~Teng, X.~Kan, and K.~Karydis, ``A data-driven hierarchical control structure for systems with uncertainty,'' in \emph{IEEE Conference on Control Technology and Applications (CCTA)}, 2020, pp. 57--63.

\bibitem{abraham2017RKexample}
I.~Abraham, G.~De~La~Torre, and T.~D. Murphey, ``Model-based control using koopman operators,'' in \emph{Robotics: Science and Systems (RSS)}, 2017.

\bibitem{bruder2019ICRA}
D.~Bruder, C.~D. Remy, and R.~Vasudevan, ``Nonlinear system identification of soft robot dynamics using koopman operator theory,'' in \emph{IEEE International Conference on Robotics and Automation (ICRA)}, 2019, pp. 6244--6250.

\bibitem{zhou2023safe}
H.~Zhou, Y.~Song, and V.~Tzoumas, ``Safe non-stochastic control of control-affine systems: An online convex optimization approach,'' \emph{IEEE Robotics and Automation Letters}, 2023.

\bibitem{nathan2018applied}
J.~Nathan~Kutz, J.~L. Proctor, and S.~L. Brunton, ``Applied koopman theory for partial differential equations and data-driven modeling of spatio-temporal systems,'' \emph{Complexity}, no.~1, p. 6010634, 2018.

\bibitem{surana2016koopman}
A.~Surana, ``Koopman operator based observer synthesis for control-affine nonlinear systems,'' in \emph{IEEE 55th Conference on Decision and Control (CDC)}, 2016, pp. 6492--6499.

\bibitem{bruder2021advantages}
D.~Bruder, X.~Fu, and R.~Vasudevan, ``Advantages of bilinear koopman realizations for the modeling and control of systems with unknown dynamics,'' \emph{IEEE Robotics and Automation Letters}, vol.~6, no.~3, pp. 4369--4376, 2021.

\bibitem{shi2021acd}
L.~Shi and K.~Karydis, ``Acd-edmd: Analytical construction for dictionaries of lifting functions in koopman operator-based nonlinear robotic systems,'' \emph{IEEE Robotics and Automation Letters}, vol.~7, no.~2, pp. 906--913, 2021.

\bibitem{bichiou2018time}
S.~Bichiou, M.~K. Bouafoura, and N.~Benhadj~Braiek, ``Time optimal control laws for bilinear systems,'' \emph{Mathematical Problems in Engineering}, no.~1, p. 5217427, 2018.

\bibitem{shi2021enhancement}
L.~Shi and K.~Karydis, ``Enhancement for robustness of koopman operator-based data-driven mobile robotic systems,'' in \emph{IEEE International Conference on Robotics and Automation (ICRA)}, 2021, pp. 2503--2510.

\bibitem{ryu2011differential}
J.-C. Ryu and S.~K. Agrawal, ``Differential flatness-based robust control of mobile robots in the presence of slip,'' \emph{The International Journal of Robotics Research}, vol.~30, no.~4, pp. 463--475, 2011.

\bibitem{andersson2019casadi}
J.~A. Andersson, J.~Gillis, G.~Horn, J.~B. Rawlings, and M.~Diehl, ``Casadi: a software framework for nonlinear optimization and optimal control,'' \emph{Mathematical Programming Computation}, vol.~11, pp. 1--36, 2019.

\bibitem{leineweber2003efficient}
D.~B. Leineweber, I.~Bauer, H.~G. Bock, and J.~P. Schl{\"o}der, ``An efficient multiple shooting based reduced sqp strategy for large-scale dynamic process optimization. part 1: theoretical aspects,'' \emph{Computers \& Chemical Engineering}, vol.~27, no.~2, pp. 157--166, 2003.

\bibitem{wachter2006implementation}
A.~W{\"a}chter and L.~T. Biegler, ``On the implementation of an interior-point filter line-search algorithm for large-scale nonlinear programming,'' \emph{Mathematical programming}, vol. 106, pp. 25--57, 2006.

\bibitem{macenski2022robot}
S.~Macenski, T.~Foote, B.~Gerkey, C.~Lalancette, and W.~Woodall, ``Robot operating system 2: Design, architecture, and uses in the wild,'' \emph{Science robotics}, vol.~7, no.~66, p. eabm6074, 2022.

\end{thebibliography}

\end{document}